\documentclass{article}
\usepackage[T1]{fontenc}
\usepackage[utf8]{inputenc}
\usepackage{ismir} 
\usepackage{amsmath,cite,url}
\usepackage{graphicx}
\usepackage{color}

\usepackage{amsfonts,amssymb}
\usepackage{subcaption}
\usepackage{xurl}
\usepackage{flushend}

\title{Conditional Diffusion as Latent Constraints for Controllable Symbolic Music Generation}

\multauthor
   {Matteo Pettenò \hspace{1cm} Alessandro Ilic Mezza \hspace{1cm} Alberto Bernardini}
   {Dipartimento di Elettronica, Informazione e Bioingegneria\\
   Politecnico di Milano, Milan, Italy\\
   {\small \texttt{matteo.petteno@mail.polimi.it,}  \texttt{alessandroilic.mezza@polimi.it,} \texttt{alberto.bernardini@polimi.it}
   }}

\def\authorname{M. Pettenò, A. I. Mezza, and A. Bernardini}

\usepackage[bookmarks=false,pdfauthor={\authorname},pdfsubject={\pdfsubject},hidelinks]{hyperref}

\begin{document}

\maketitle

\begin{abstract}
Recent advances in latent diffusion models have demonstrated state-of-the-art performance in high-dimensional time-series data synthesis while providing flexible control through conditioning and guidance. However, existing methodologies primarily rely on musical context or natural language as the main modality of interacting with the generative process, which may not be ideal for expert users who seek precise fader-like control over specific musical attributes. In this work, we explore the application of denoising diffusion processes as plug-and-play latent constraints for unconditional symbolic music generation models. We focus on a framework that leverages a library of small conditional diffusion models operating as implicit probabilistic priors on the latents of a frozen unconditional backbone. While previous studies have explored domain-specific use cases, this work, to the best of our knowledge, is the first to demonstrate the versatility of such an approach across a diverse array of musical attributes, such as note density, pitch range, contour, and rhythm complexity. Our experiments show that diffusion-driven constraints outperform traditional attribute regularization and other latent constraints architectures, achieving significantly stronger correlations between target and generated attributes while maintaining high perceptual quality and diversity.
\end{abstract}

\section{Introduction}
\label{sec:introduction}
Latent Constraints (LC) \cite{engel2018latent} refer to a set of techniques for generating conditionally from unconditional generative models. Traditional methods for enforcing user-defined constraints during model training, such as using attribute-regularization losses or training on curated subsets, require extensive labeled data, large amounts of computational power, and time-consuming hyperparameter tuning. The cost of retraining becomes increasingly prohibitive as the number of conditioning variables grows, 
especially when expert users need the flexibility to choose from a range of different attributes at any given time, as is the case in symbolic music generation and computer-assisted music composition.

Deep latent-variable models like GANs and VAEs learn to generate diverse outputs by sampling from a structured latent space. By exploiting this property, LC provides a principled framework for endowing pre-trained unsupervised models with post-hoc conditional generation capabilities. This is achieved either \textit{explicitly}, by optimizing a new model that imposes the desired behavior onto latent representations \cite{engel2018latent}, or \textit{implicitly}, by training a small personalized model to generate only from regions of the latent space  \cite{midime}. In this way, at inference time, LC models yield latents that, once decoded, result in outputs with the desired attributes. LC is also closely related to \textit{latent translation}~\cite{tian2019latent}, which introduces neural networks that bridge multimodal representations of different pre-trained generative models, conditioning on the respective domain labels.

Latent diffusion \cite{rombach2022high} can be thought of as a class of LC. In Latent Diffusion Models (LDMs), the pre-trained autoencoder is typically understood as a way to compress the data into a lower-dimensional space where reverse diffusion is computationally feasible. 
By conditioning the denoising process, though, it is possible to steer the decoder to generate outputs with desired characteristics by feeding it inputs that lie in certain regions of the latent space associated with the desired attributes of the output, just like in existing LC methods.

Diffusion-based symbolic music generation models have been conditioned on various inputs, including text prompts \cite{wu2022exploring, jajoria2024text}, musical context \cite{mittalSymbolicMusicGeneration2021}, accompaniment \cite{pasini2024bass}, chords \cite{li2023melodydiffusion, polyffusion2023}, and rhythmic textures \cite{polyffusion2023}.
Recent methods, such as MelodyDiffusion~\cite{li2023melodydiffusion} and Polyffusion~\cite{polyffusion2023}, apply conditional denoising on piano roll representations, treating them as image-like data. As a result, sequence modeling is tightly coupled with control, making it infeasible to seamlessly substitute one conditioning signal for another without retraining the note-generation process.
Notably, this also applies to many recent controllable generation methods that are not based on diffusion~\cite{kawai2020attributes, MuseMorphose, malandro2024}.

Against this backdrop, diffusion-driven LC offer a particularly compelling approach for achieving modular fader-like control \cite{lample2017fader, tan20music} over multiple musical attributes with an otherwise unconditional model, allowing users to manipulate different musical features along continuous axes through a range of attribute-specific LDMs.

LDM specifications depend on the base unconditional model; Denoising Diffusion Probabilistic Models (DDPM) \cite{ho2020ddpm} and Denoising Diffusion Implicit Models (DDIM) \cite{song2021denoising} operate on continuous latent spaces, whereas Discrete DDPM \cite{austin2021structured} operate on tokenized representations. 
In particular, discrete diffusion models have recently shown promising results for symbolic music generation \cite{lv2023getmusic, plasser2023discrete, zhang2024discrete}.
Prior work also explored latent diffusion for specific domains, such as emotion-controlled symbolic music generation either by learning from emotion-labeled data \cite{zhang2023fast} or by relying on emotion classifier guidance \cite{zhang2024emotionally}.
Post-hoc control over black-box music rules has been tackled in \cite{huang2024symbolic} by means of \textit{stochastic control guidance}, which, inspired by control theory, entails sampling several realizations of the next denoising step and selecting the one most compliant with the rule.

In this work, by looking at latent diffusion through the lens of LC, we study LDMs as plug-and-play conditioning modules. Thus, we keep the base generative model fixed and develop a library of diffusion-driven LC models (``LC-Diff'') trained on a range of non-differentiable and possibly continuous musical attributes, including contour, note density, pitch range, and rhythm complexity.

We show that, compared to attribute-regularized VAEs \cite{patiAttributebasedRegularizationLatent2021, mezzaLatentRhythmComplexity2023} and other LC architectures \cite{tian2019latent}, LC-Diff improves fidelity (measured by Fr{\'e}chet Music Distance \cite{retkowski2024frechet}) and controllability (measured by the correlation between desired attributes and those of generated samples) across all attributes considered in the present study.

\section{Diffusion as Latent Constraints}\label{sec:diffusion}

Let $z \sim p(z|x)$ be the latent representation of an input sequence $x$ with $N$ tokens and attribute $a\in\mathbb{R}$.
Diffusion models employ a Markov chain to progressively corrupt input data with Gaussian noise and learn to reverse the process. 
Forward (latent) diffusion begins with the representation $z_0 = z$ and gradually adds noise following a schedule $\beta_t$, with $t = 1, \dots, T$. At each step, Gaussian noise is introduced according to
\begin{equation}
    q(z_t | z_{t-1}) = \mathcal{N}(z_t; \sqrt{1 - \beta_t} z_{t-1},\, \beta_t \mathbf{I}).
\end{equation}
The LC-Diff reverse diffusion process aims to navigate the latent space of a pre-trained generative model by tracing a trajectory conditional on the target attribute starting from a noise sample $z_T\sim\mathcal{N}(\mathbf{0}, \mathbf{I})$.
A denoising function $\epsilon_\theta$ is trained to predict the additive noise at each step. $\epsilon_\theta$ is thus conditioned on $a$ and a time variable $\xi_t$ that can be either the diffusion step $\xi_t=t$ \cite{ho2020ddpm} or the continuous noise level $\xi_t=\sqrt{\bar{\alpha}_t}$ \cite{chen2021wavegrad}, where $\bar{\alpha}_t = \prod^{t}_{i=1}(1-\beta_i)$.

By sampling the conditional distribution of $z_t$ at an arbitrary timestep in closed form
\begin{equation}
    q(z_t | z_0) = \mathcal{N}(z_t; \sqrt{\bar{\alpha}_t} z_0,\, (1 - \bar{\alpha}_t) \mathbf{I}),
\end{equation}
it is possible to efficiently train $\epsilon_\theta$ by optimizing random terms of the following objective \cite{ho2020ddpm}:
\begin{equation}\label{eq:latent_diffusion_loss}
    \mathcal{L} = \mathbb{E}_{z_0\sim p(z|x), \epsilon\sim\mathcal{N}(\mathbf{0},\mathbf{I}), t} \left[ \|\epsilon - \epsilon_\theta(z_t, \xi_t, a) \|^2 \right],
\end{equation}
where $\xi_t$ is sampled from either
$\mathcal{U}(\{1,\dots, T\})$ \cite{ho2020ddpm} or 
$\mathcal{U}( \sqrt{\bar{\alpha}_{t-1}}, \sqrt{\bar{\alpha}_t})$ \cite{chen2021wavegrad}.



\begin{figure}
    \centering
    \includegraphics[width=0.75\linewidth,alt={
    Block diagram depicting the architecture of the conditioning networks of LC-Diff. The left network processes the noise level. The right network processes the musical attribute.}]{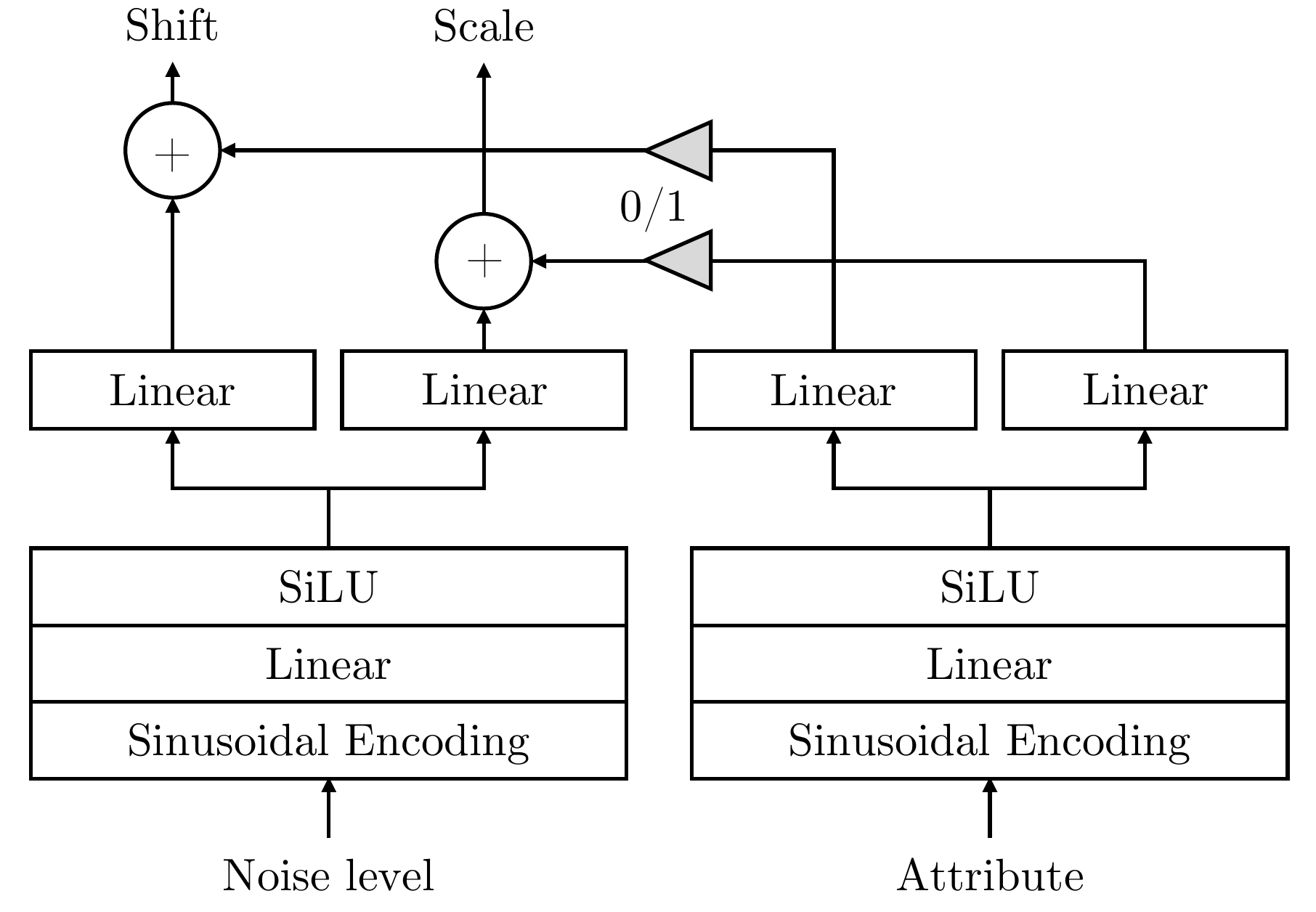}
    \caption{LC-Diff conditioning networks.}
    \label{fig:conditioning_network}
\bigskip
    \centering
    \includegraphics[width=0.77\linewidth,alt={
    Block diagram depicting the denoiser network of the LC-Diff model.}]{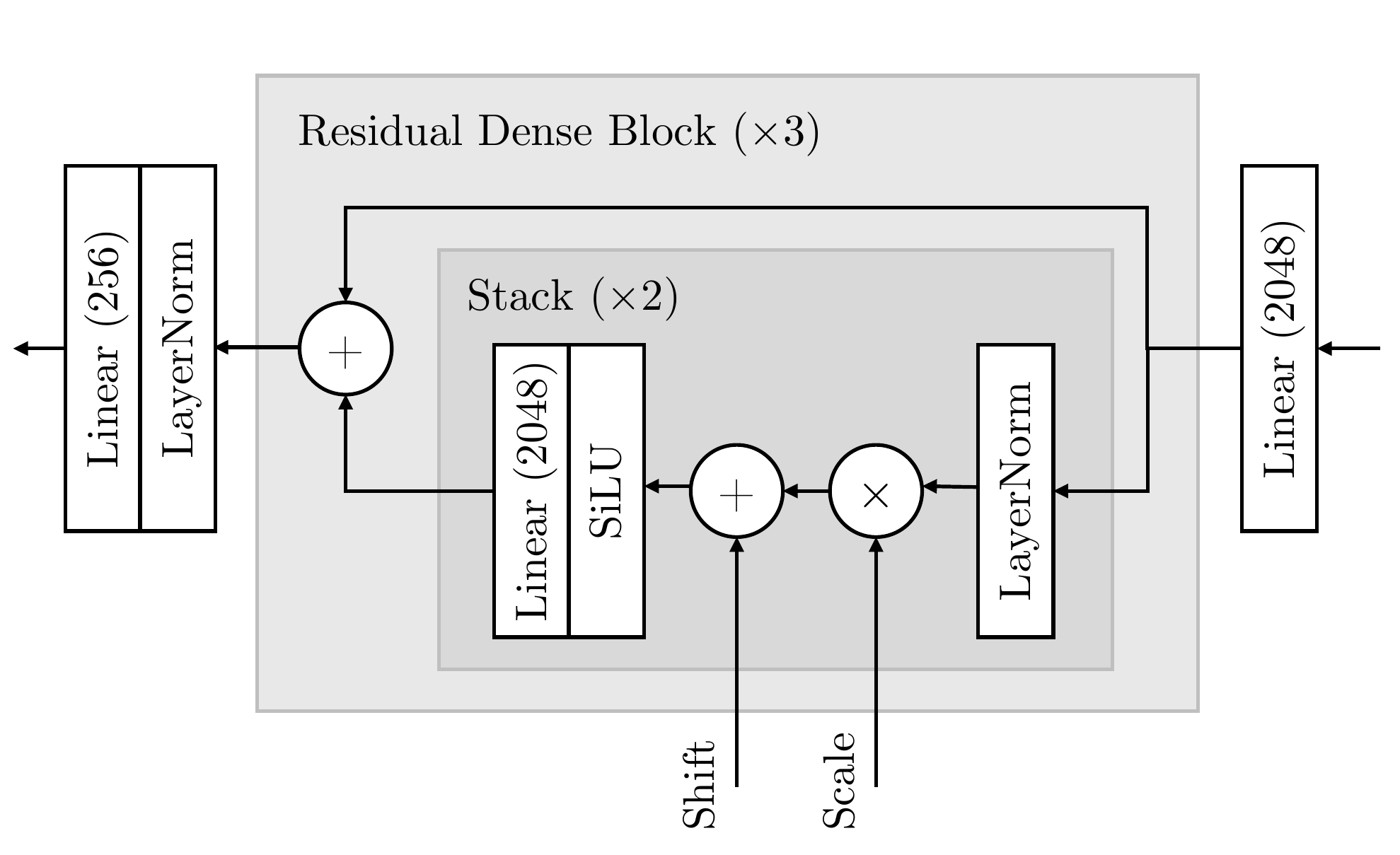}
    \caption{LC-Diff denoiser architecture.}
    \label{fig:denoiser}
\end{figure}

\subsection{Sampling} 
Different sampling strategies have been explored in the literature. At inference time, DDPMs \cite{ho2020ddpm} involve a stochastic Markov process where, at each intermediate step, a small amount of Gaussian noise is added back in to encourage diversity in the generated samples. However, following a stochastic trajectory typically requires a large number of steps, slowing down the sampling process.
DDIMs \cite{song2021denoising} differ from DDPMs by
making the process deterministic. With this class of models, forward diffusion is reversed by
\begin{equation}
    z_{t-1} = \sqrt{\bar{\alpha}_{t-1}} f(z_t, \xi_t, a) + g(z_t, \xi_t, a),
\end{equation}
where
\begin{equation}
    f(z_t, \xi_t, a) = \frac{z_t - \sqrt{1 - \bar{\alpha}_t} {\epsilon}_\theta(z_t, \xi_t, a)}{\sqrt{\bar{\alpha}_t}}
\end{equation}
attempts to directly estimate $z_0$ from the current noisy latent $z_t$, while
\begin{equation}
    g(z_t, \xi_t, a) = \sqrt{1 - \bar{\alpha}_{t-1}} {\epsilon}_\theta(z_t, \xi_t, a)
\end{equation}
ensures that the trajectory toward $z_0$ follows the direction pointing to $z_t$.
The deterministic nature of DDIM allows to skip intermediate denoising steps and perform only $T_\mathrm{s}\ll T$ iterations of the reverse process, thereby enabling faster inference.

\subsection{Conditioning}
We aim to condition $\epsilon_\theta$ on continuous musical attributes $a\in\mathbb{R}$.
Similarly, Chen et al.~\cite{chen2021wavegrad} found it beneficial to condition on the noise level instead of the discrete diffusion step, which \cite{mittalSymbolicMusicGeneration2021} later adopted for LDM-based symbolic music generation. Thus, we are left with two continuous conditioning signals to be passed onto the diffusion model. 

We inject $a$ and $\sqrt{\bar{\alpha}_t}$ into $\epsilon_\theta$ through dedicated networks (see Figure~\ref{fig:conditioning_network}). First, we apply Sinusoidal Encoding (SE) based on Transformer positional embeddings \cite{attention_is_all_you_need}
\begin{equation}\label{eq:se}
\Gamma(u) = \big[ \sin \left( \omega_i(u) \right), \cos \left( \omega_i(u) \right) \big]_{i=0}^{d/2}
\end{equation}
where $\omega_i(u) = s \frac{u}{b^{2i/d}}$ \cite{mittalSymbolicMusicGeneration2021}, with  $d\in\mathbb{N}$ the (even) dimensionality of the embedding, $b\in\mathbb{R}$ the base frequency, and $s\in\mathbb{R}$ a frequency scaling hyperparameter. The resulting SE features are passed through a linear layer with SiLU activations. 
Finally, we employ Feature-wise Linear Modulation (FiLM) \cite{perez2018film}, where two fully-connected layers yield \textit{shifts} and \textit{scales}, respectively, that modulate the activations of the denoiser (see Figure~\ref{fig:denoiser}).
The two conditioning branches run in parallel. This is equivalent to learning a single affine transformation, where scale and shift are the sum of FiLM outputs from the attribute and noise level conditioning networks.

To enhance controllability over the generated samples, we also apply Classifier-Free Guidance (CFG) \cite{hoClassifierFreeDiffusionGuidance2022} to the noise prediction:
\begin{equation}
    \hat{\epsilon}_\theta(z_t, \xi_t, a) = (1 + w) \epsilon_\theta(z_t, \xi_t, a) - w \epsilon_\theta(z_t, \xi_t),
\end{equation}
where $\epsilon_\theta(z_t, \xi_t)$ is the unconditional noise prediction and $w\in\mathbb{R}_{\ge 0}$ is the guidance scale. To make CFG effective, the model must learn to predict noise both with and without attribute conditioning. We achieve this through \textit{conditioning dropout} (depicted as $0/1$ in Figure~\ref{fig:conditioning_network}), i.e., setting the outputs of the attribute conditioning network to zero with a certain probability when evaluating \eqref{eq:latent_diffusion_loss}.

\section{Evaluation}\label{sec:evaluation}

\begin{table}[t]
    \centering
    \resizebox{\linewidth}{!}{
    \begin{tabular}{|c|c|c|c|c|}
        \hline
         (\%) & \textbf{Contour} & \textbf{Note Density} & \textbf{Pitch Range} & \textbf{Complexity} \\
        \hline
         NM & $63.36$ & $76.63$ & $46.70$ &  $48.67$\\
         P\&L & $37.44$ & $10.11$ & $0.41$ &  $50.59$\\
         \hline
         LC-VAE-A & $60.88$ & $97.56$ & $34.69$ &  $33.32$\\
         LC-VAE-SE & $52.02$ & $97.33$ & $36.84$ & $52.00$ \\
         \hline
         LC-Diff & $\mathbf{85.60}$ & $\mathbf{98.59}$ & $\mathbf{80.97}$ & $\mathbf{94.93}$\\
         \hline         
    \end{tabular}
    }
    \caption{Pearson Correlation Coefficient (PCC) between target and decoded attributes.}
    \label{tab:pearson_correlation}
    \vspace{-0.33em}
\end{table}

\subsection{Dataset}\label{ssec:dataset}

The models are designed to learn pitch sequence representations from four-bar monophonic melodies. We construct a large-scale dataset comprising melodies extracted from 176,581 MIDI files from the Lakh MIDI Dataset~\cite{raffelLearningBasedMethodsComparing2016a}.\footnote{C. Raffel, 2016, ``The Lakh MIDI Dataset v0.1.'' [Online]. Available: \url{https://colinraffel.com/projects/lmd}}

First, we assess whether each MIDI file contains time signature changes. If any are found, we segment the file and retain only sections with a $4/4$ time signature. Each MIDI event is then quantized to the nearest sixteenth note.

A melody is defined as a sequence of pitches within the standard 88-key piano range, played by an instrument mapped to a valid MIDI program. A melody is considered complete when a full measure of silence occurs. We extract only melodies spanning at least four bars and comprising at least three distinct pitches. If multiple notes sound simultaneously, we follow the approach proposed in~\cite{robertsHierarchicalLatentVector2018} and select only the highest-pitched note to ensure monophonic sequences. Subsequently, four-bar segments are extracted using a stride of one bar.

For each melody thus extracted, we compute $13$ musical attributes, including those outlined in Section~\ref{ssec:attributes}. 

Melodies are encoded as sequences of $N=64$ integers in $\mathbb{P}=\{0, \dots, 129\}$, where each element represents either a MIDI note number ($0$-$127$) or one of two special tokens: \textit{note off} ($128$) and \textit{note hold} ($129$). The dataset is divided into training, validation, and test sets, with training data augmented through transposition by a randomly selected number of semitones within a range of $\pm1$ octave.
The final dataset, consisting of $10,\!126,\!676$ unique melodies, is publicly available.\footnote{M.~Pettenò, Aug.~2024, ``4 Bars Monophonic Melodies Dataset (Pitch Sequence),'' Zenodo, doi: \url{https://doi.org/10.5281/zenodo.13369389}}

\begin{table}[t]
    \centering
    \resizebox{\linewidth}{!}{
    \begin{tabular}{|c|c|c|c|c|}
        \hline
         & \textbf{Contour} & \textbf{Note Density} & \textbf{Pitch Range} & \textbf{Complexity} \\
        \hline
        Uncond.~VAE & \multicolumn{4}{c|}{$41.44$} \\
        \hline
         NM & $35.506$ & $58.436$ & $30.833$ & $47.61$ \\
         P\&L & $49.698$ & $67.836$ & $40.657$ & $87.80$ \\
         \hline
         LC-VAE-A & $30.197$ & $29.450$ & $\mathbf{30.257}$ & $32.435$ \\
         LC-VAE-SE & $29.161$ & $30.124$ & $31.274$ & $30.166$ \\
         \hline
         LC-Diff & $\mathbf{19.299}$ & $\mathbf{20.559}$ & $31.695$ & $\mathbf{17.51}$\\
         \hline
    \end{tabular}
    }%
    \caption{Fr{\'e}chet Music Distance \cite{retkowski2024frechet}.}
    \label{tab:frechet_music_distance}
\end{table}

\subsection{Musical Attributes}
\label{ssec:attributes}


As previously done in~\cite{patiAttributebasedRegularizationLatent2021}, we focus on four musical attributes:
(i)~\textbf{Contour}, which quantifies the melodic movement in a sequence, measured by averaging the pitch differences between consecutive notes; 
(ii)~\textbf{Note Density}, defined as the ratio between the number of notes in the melody and the sequence length. It takes values in $[0, 1]$; 
(iii)~\textbf{Pitch Range}, defined as the difference between the highest and lowest MIDI pitch values in the sequence, normalized by the range of an $88$-key piano. It takes values in $\left[0, \frac{127}{88}\right]$, where values above one indicate a range exceeding \mbox{A0--C8};
(iv)~\textbf{Rhythm Complexity}, evaluated using Toussaint’s metrical complexity measure~\cite{toussaintMathematicalAnalysisAfrican2002},
corrected for the total number of notes in the sequence~\cite{mezzaLatentRhythmComplexity2023}. By definition, it takes on discrete values.

\begin{figure*}[t]
\centering
    \begin{subfigure}{4.2cm}
        \centering
        \includegraphics[width=\linewidth,alt={A scatter plot showing a cloud of points, where the x-axis represents the desired Contour attributes and the y-axis represents the Contour attributes of the corresponding sequences decoded by the NM model. A best-fit line illustrates the linear regression between them.}]{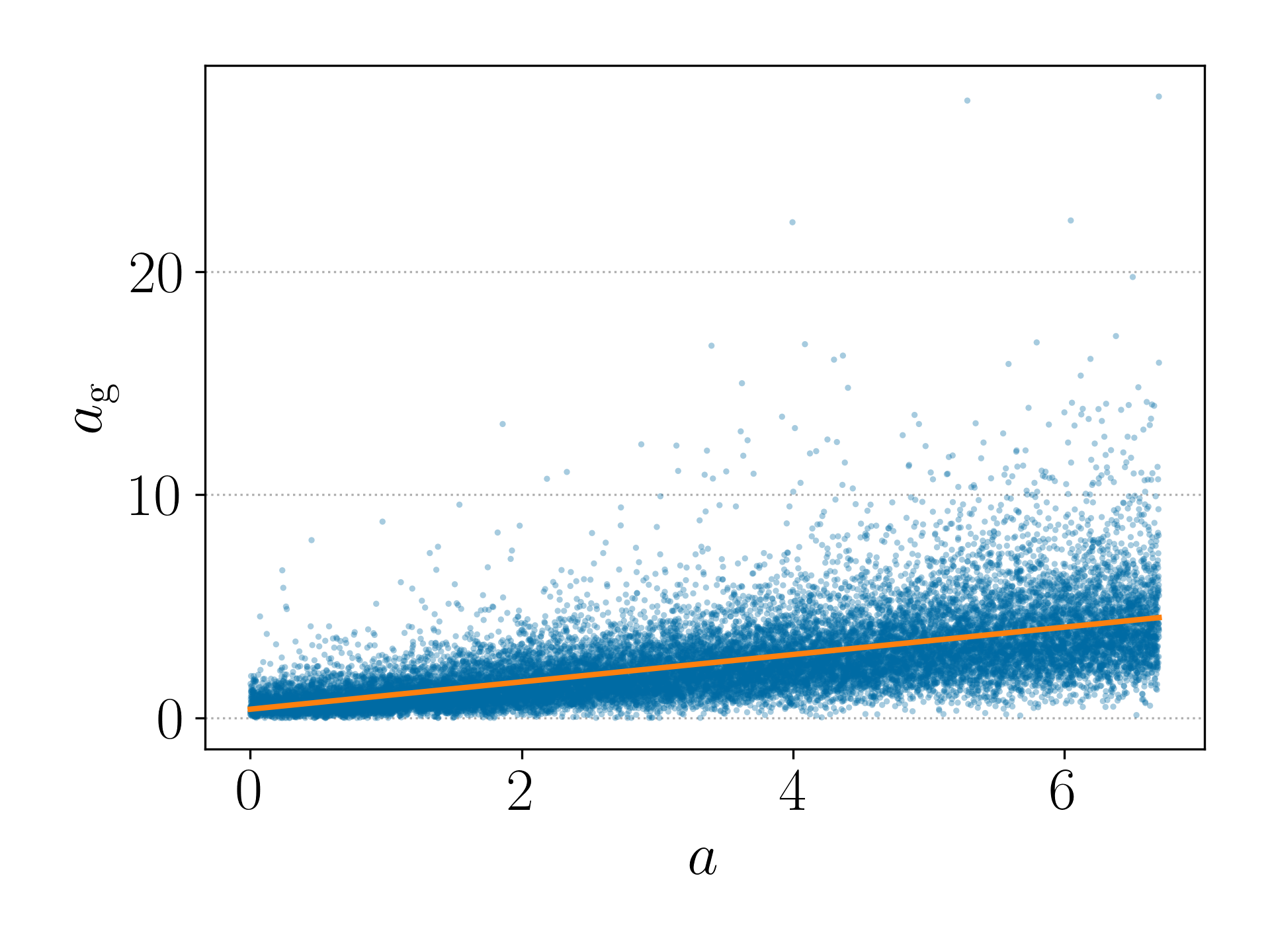}
        \caption{NM}
    \end{subfigure}
    \begin{subfigure}{4.2cm}
        \centering
        \includegraphics[width=\linewidth,alt={A scatter plot showing a cloud of points, where the x-axis represents the desired Contour attributes and the y-axis represents the Contour attributes of the corresponding sequences decoded by the P\&L model. A best-fit line illustrates the linear regression between them.}]{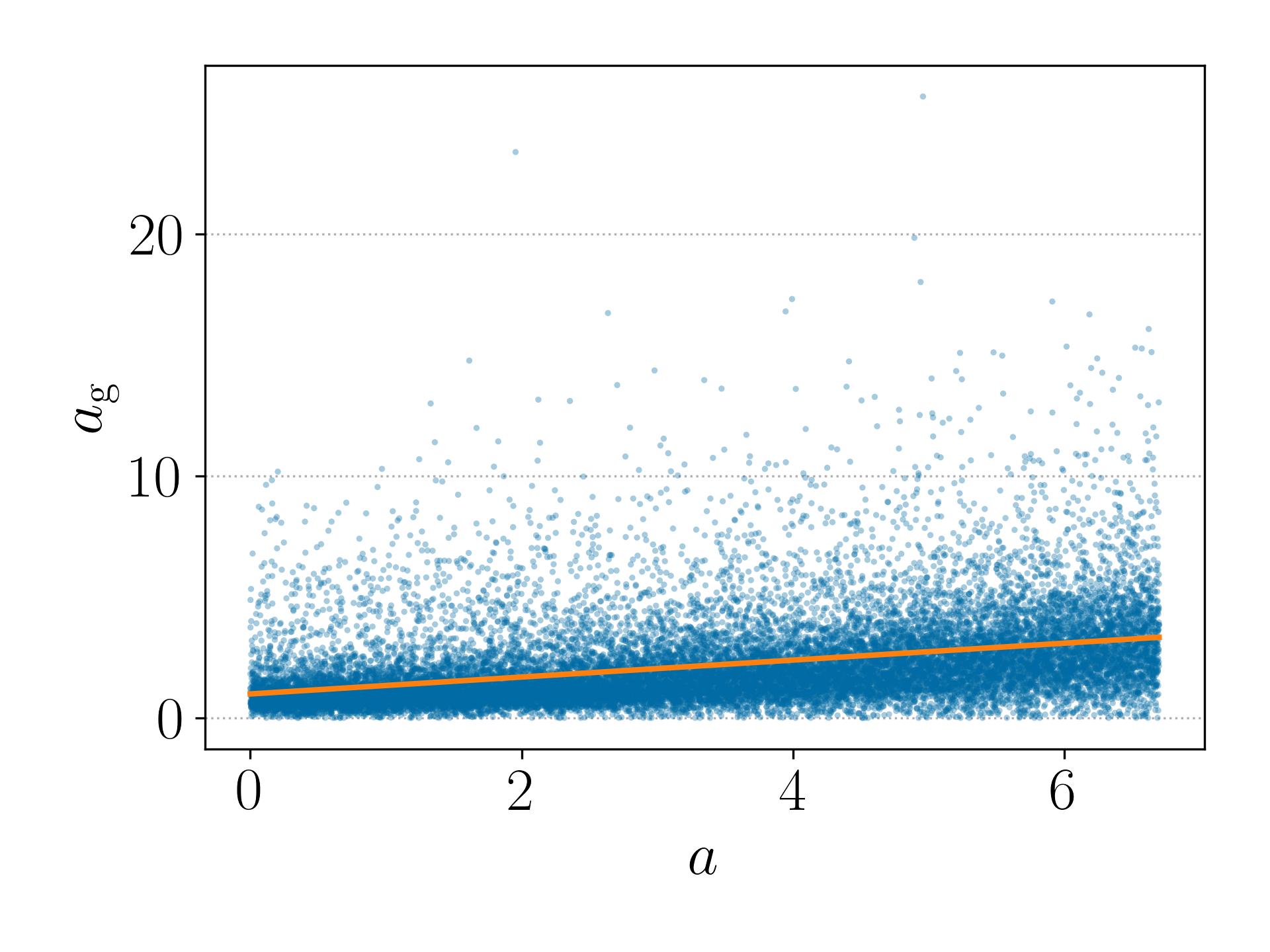}
        \caption{P\&L}
    \end{subfigure}
    \begin{subfigure}{4.2cm}
        \centering
        \includegraphics[width=\linewidth,alt={A scatter plot showing a cloud of points, where the x-axis represents the desired Contour attributes and the y-axis represents the Contour attributes of the corresponding sequences decoded by the LC-VAE-A model. A best-fit line illustrates the linear regression between them.}]{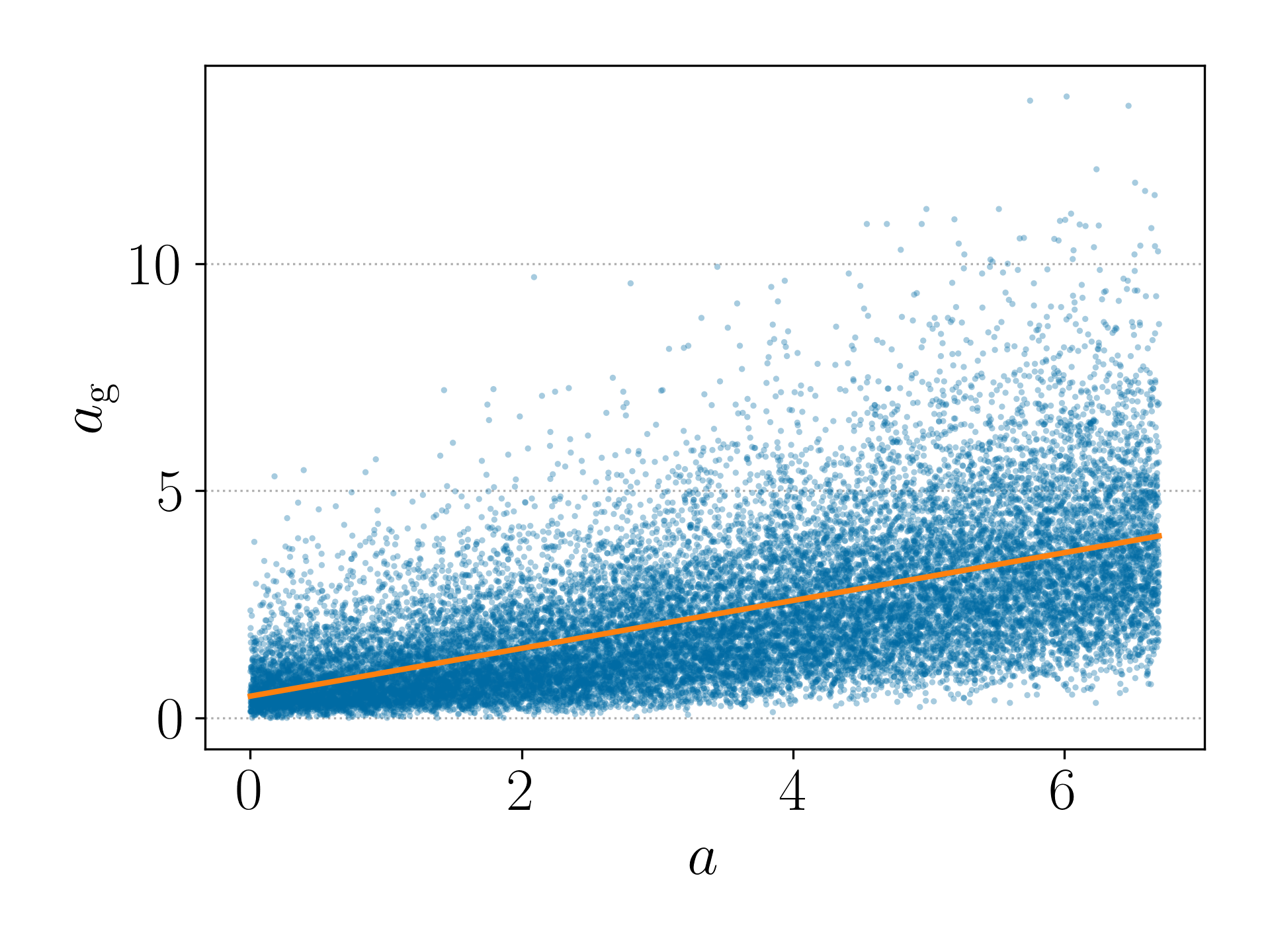}
        \caption{LC-VAE-A}
    \end{subfigure}
    \begin{subfigure}{4.2cm}
        \centering
        \includegraphics[width=\linewidth,alt={A scatter plot showing a cloud of points, where the x-axis represents the desired Contour attributes and the y-axis represents the Contour attributes of the corresponding sequences decoded by the LC-VAE-SE model. A best-fit line illustrates the linear regression between them.}]{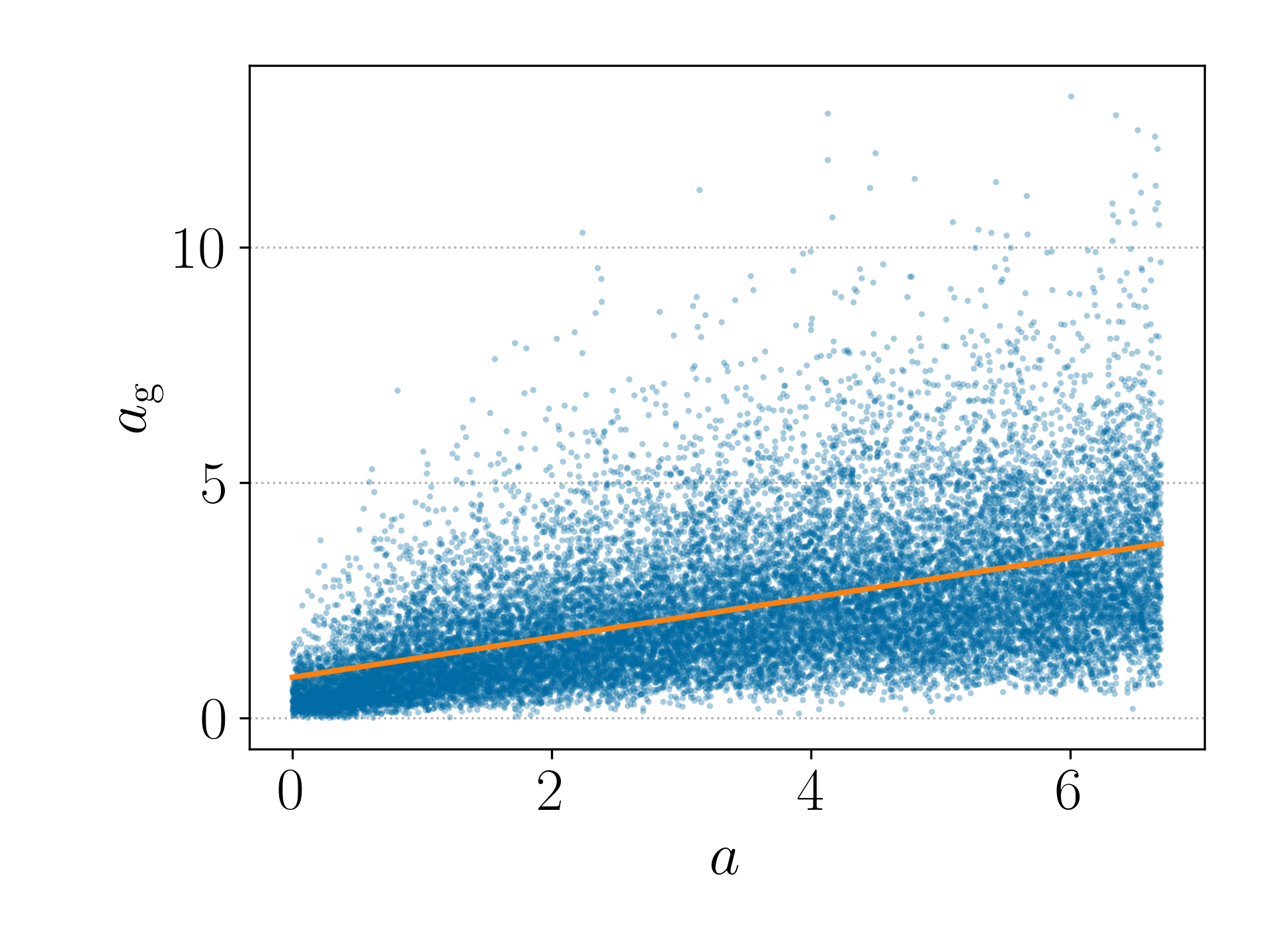}
         \caption{LC-VAE-SE}
    \end{subfigure}
    \vspace{-0.33em}
    \caption{Regression plots comparing target and decoded Contour attributes across baseline methods.}
    \label{fig:reg-plots-contour}
    \vspace{-0.5em}
\end{figure*}
\begin{figure}
        \centering
        \includegraphics[width=4.8cm,alt={A scatter plot showing a cloud of points, where the x-axis represents the desired Contour attributes and the y-axis represents the Contour attributes of the corresponding sequences decoded by the LC-Diff model. A best-fit line illustrates the linear regression between them.}]{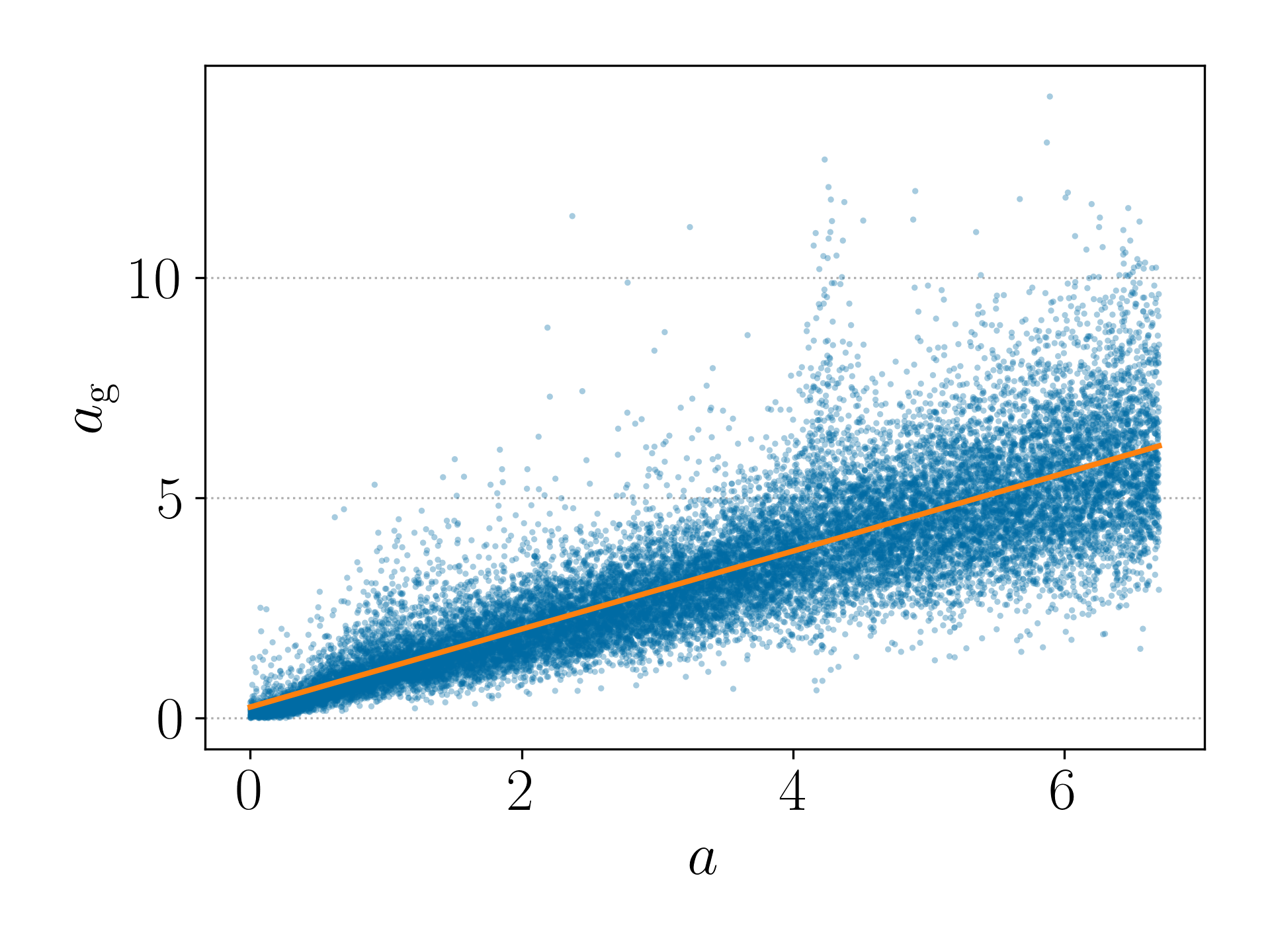}
    \vspace{-1em}
    \caption{Regression plot comparing target and decoded Contour attributes using LC-Diff.}
    \label{fig:reg-plots-contour-lc-diff}
    \vspace{-0.5em}
\end{figure}

\subsection{Unconditional Generative Model}
\label{ssec:unconditional_model}
As base unconditional model, we implement a $\beta$-VAE~\cite{higgins2017beta} based on MusicVAE~\cite{robertsHierarchicalLatentVector2018}. 
This model, previously used in LDM-based symbolic music generation~\cite{mittalSymbolicMusicGeneration2021}, also enables direct comparison with existing attribute-regularized VAEs (AR-VAEs) employing the same architecture~\cite{patiAttributebasedRegularizationLatent2021, mezzaLatentRhythmComplexity2023} (see Section~\ref{ssec:baselines}).

The encoder $p_\psi(z|x)$ consists of a two-layer bidirectional LSTM network fed with four-bar pitch sequence representations (see Section~\ref{ssec:dataset}), followed by two linear layers parameterizing the latent posterior. The hierarchical decoder $q_\phi(x|z)$ features two unidirectional LSTMs, with the bottom-level network autoregressively estimating the distribution over the sequence values via a softmax nonlinearity~\cite{robertsHierarchicalLatentVector2018}.
As such, each pitch sequence $x\in\mathbb{P}^{N}$ is first mapped onto a single latent code $z\in\mathbb{R}^M$, $M=256$, and since decoding amounts to a next token prediction task, the standard $\beta$-VAE objective \cite{higgins2017beta}
\begin{equation}
\label{eq:vae_objective} 
\textstyle
    \mathcal{L}_{\text{VAE}} =  -\mathbb{E}_{p_{\psi}(z|x)}\left[\log q_{\phi}(x|z)\right] + \beta D_{\text{KL}} \left[ p_{\psi}(z|x)\!\parallel\!p(z) \right],
\end{equation}
is implemented using cross-entropy as reconstruction loss.

The unconditional model is trained for $40,\!000$ iterations on a single NVIDIA Titan RTX GPU with a batch size of $512$. The objective \eqref{eq:vae_objective} is minimized using Adam, and the learning rate is decreased exponentially from $10^{-3}$ to $10^{-5}$ with a rate of $0.9999$. The hyperparameter $\beta$ is annealed exponentially from $0$ to $10^{-3}$, which encourages  the model to prioritize accurate sequence reconstruction during the early part of the training. Similarly to~\cite{robertsHierarchicalLatentVector2018}, we apply teacher forcing within the bottom-level decoder with a probability following a logistic schedule.


\subsection{Conditional Diffusion Model}
With latent codes being vectors in $\mathbb{R}^M$, we implement a DDIM model with a fully-connected denoiser network.\footnote{Source code and audio examples are available at \url{https://mpetteno.github.io/controllable-latent-diffusion/}} Shown in Figure~\ref{fig:denoiser}, the denoiser comprises an input layer with $2048$ linear units, followed by three dense residual blocks. Each residual block comprises two stacks of LayerNorm,
feature-wise modulation (responsible for joint attribute and time conditioning), SiLU, and a linear layer, plus a residual connection that shortcuts the input and output of the block. Finally, the output is linearly projected back onto $\mathbb{R}^M$.

We set the SE dimensionality to $d=128$. The attribute and noise level conditioning networks have $512$ and $2048$ units in the first linear layer and FiLM layers, respectively.

In the forward process, $\beta_t$ follows a linear schedule from $10^{-6}$ to $10^{-2}$ over $T=1000$ steps. Conversely, the number of sampling steps is set to $T_\mathrm{s}=100$.

We train the model with an attribute conditioning dropout probability of $20\%$. We then apply CFG with a guidance scale of $w=3.0$~\cite{hoClassifierFreeDiffusionGuidance2022}.
In our experiments, CFG proved fundamental to achieve attribute regularization.

The resulting denoiser network has $43.1$ million parameters, and converges in just about $20$ training epochs, half the iterations required by the unconditional model.

\begin{figure*}[t]
\centering
        \begin{subfigure}{4.2cm}
            \centering
            \includegraphics[width=\linewidth,alt={A scatter plot showing a cloud of points, where the x-axis represents the desired Rhythm Complexity attributes and the y-axis represents the Rhythm Complexity attributes of the corresponding sequences decoded by the NM model. A best-fit line illustrates the linear regression between them.}]{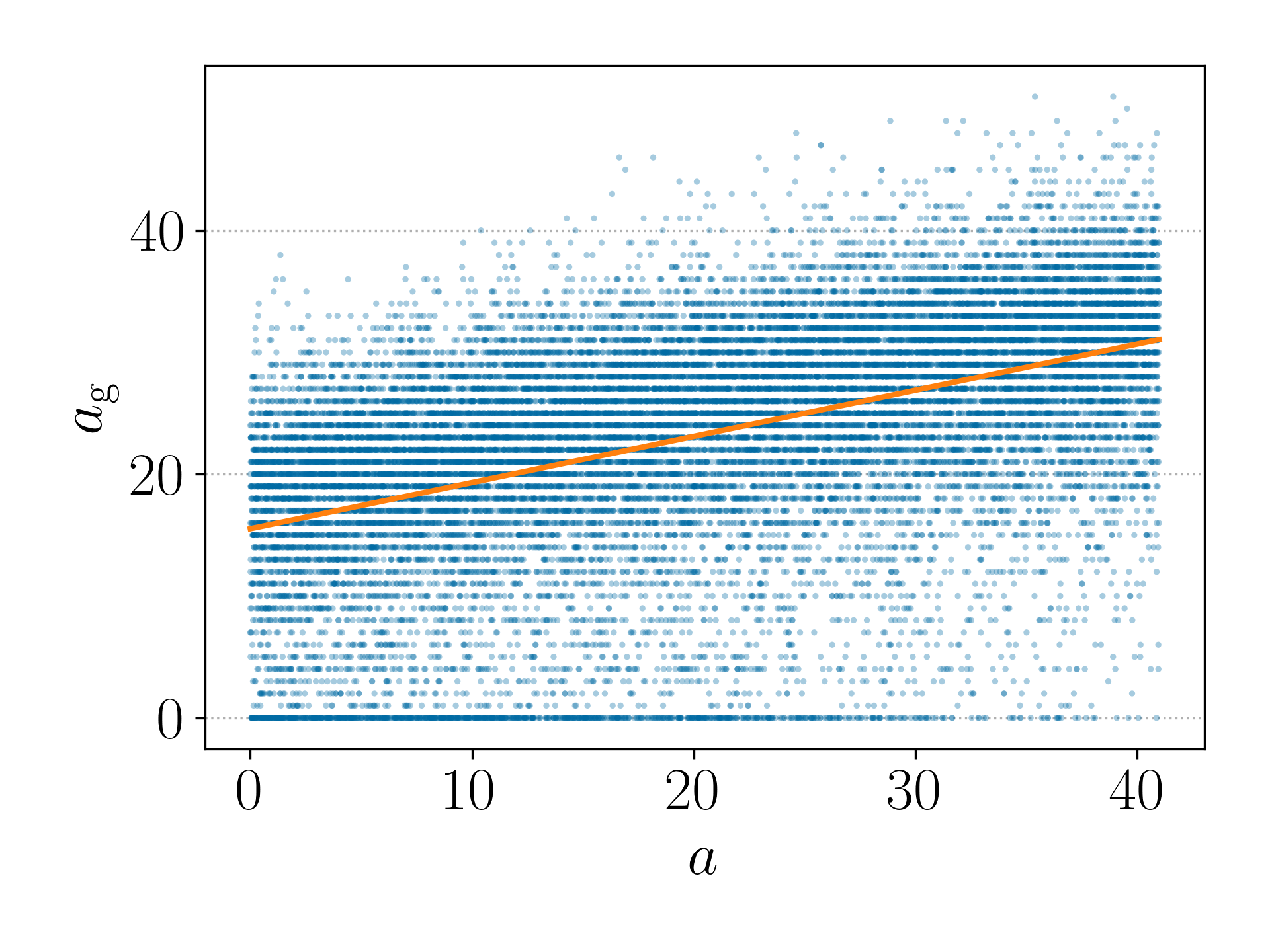}
            \caption{NM}
        \end{subfigure}
        \begin{subfigure}{4.2cm}
            \centering
             \includegraphics[width=\linewidth,alt={A scatter plot showing a cloud of points, where the x-axis represents the desired Rhythm Complexity attributes and the y-axis represents the Rhythm Complexity attributes of the corresponding sequences decoded by the P\&L model. A best-fit line illustrates the linear regression between them.}]{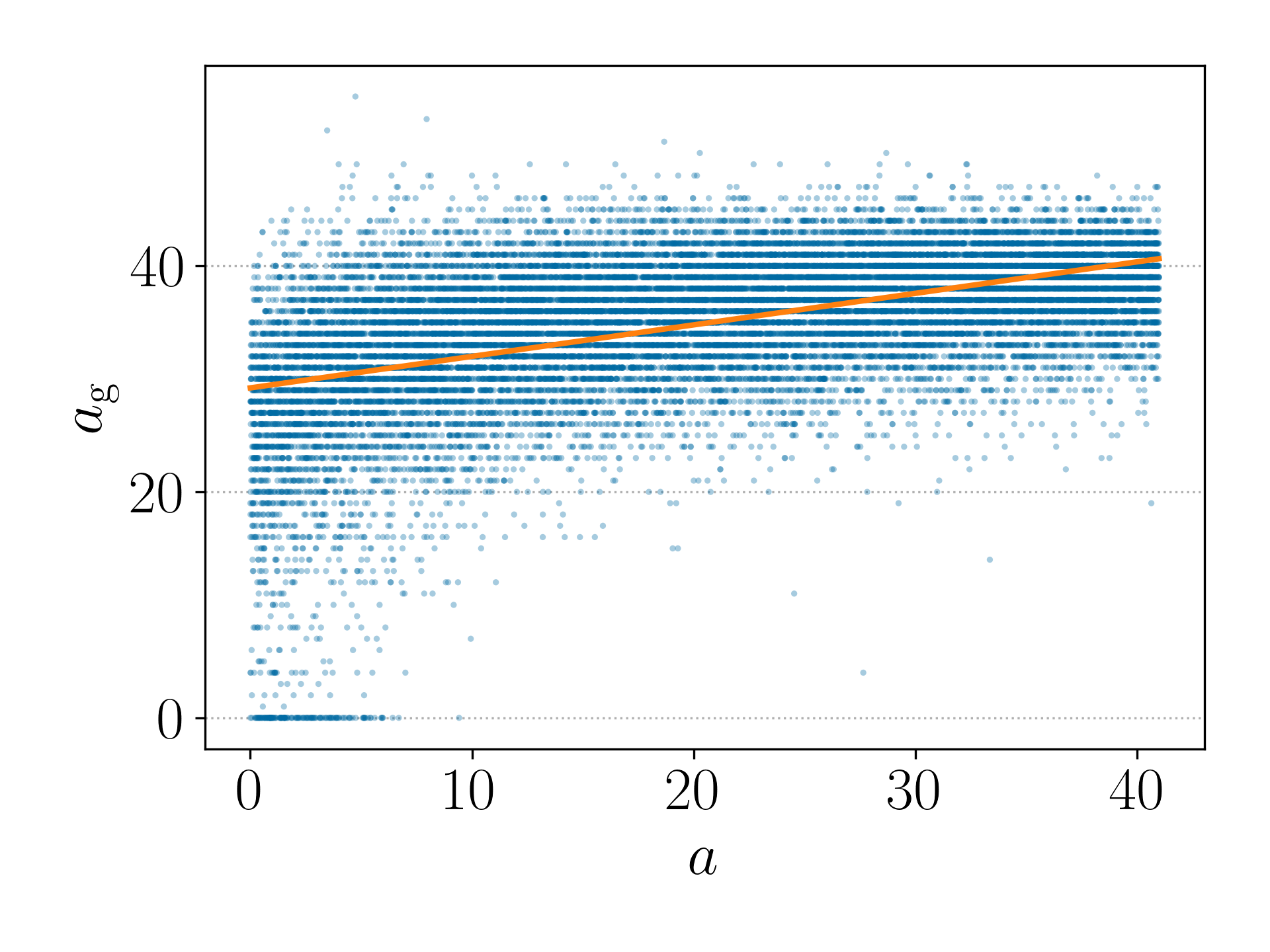}
            \caption{P\&L}
        \end{subfigure}
        \begin{subfigure}{4.2cm}
            \centering
            \includegraphics[width=\linewidth,alt={A scatter plot showing a cloud of points, where the x-axis represents the desired Rhythm Complexity attributes and the y-axis represents the Rhythm Complexity attributes of the corresponding sequences decoded by the LC-VAE-A model. A best-fit line illustrates the linear regression between them.}]{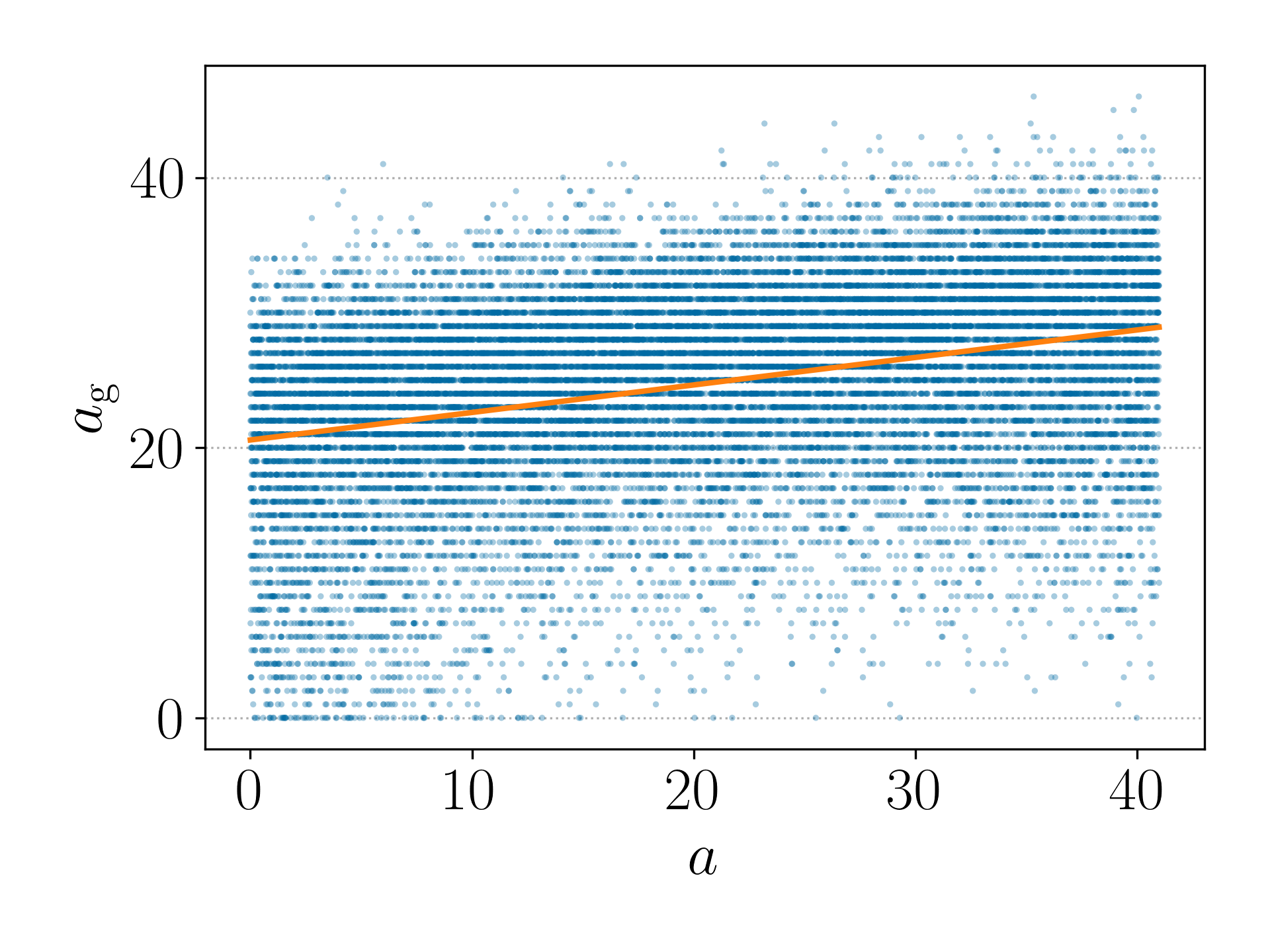}
            \caption{LC-VAE-A}
        \end{subfigure}
        \begin{subfigure}{4.2cm}
         \centering
         \includegraphics[width=\linewidth,alt={A scatter plot showing a cloud of points, where the x-axis represents the desired Rhythm Complexity attributes and the y-axis represents the Rhythm Complexity attributes of the corresponding sequences decoded by the LC-VAE-SE model. A best-fit line illustrates the linear regression between them.}]{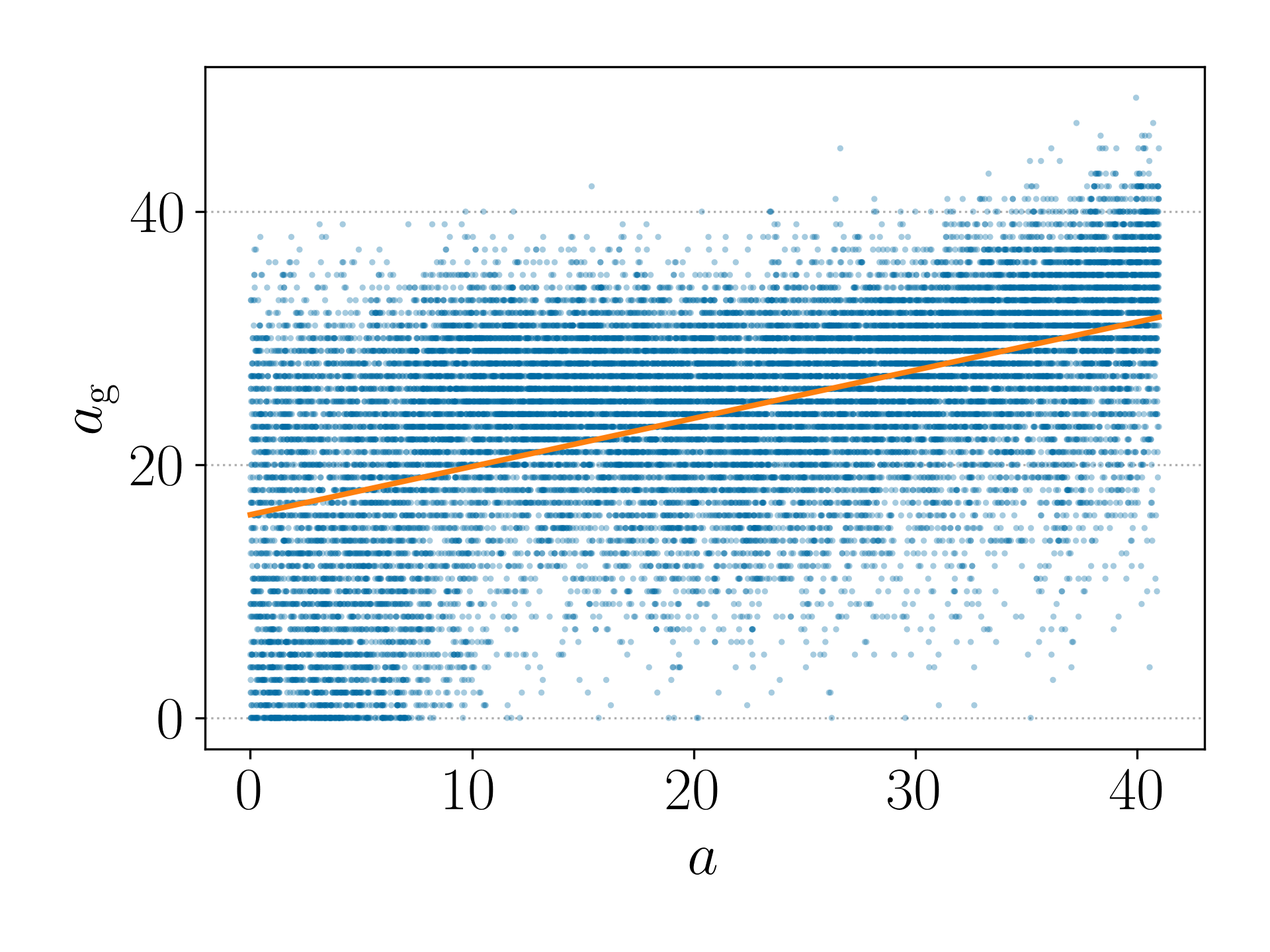}
         \caption{LC-VAE-SE}
        \end{subfigure}
    \vspace{-0.33em}
    \caption{Regression plots comparing target and decoded Rhythm Complexity attributes across baseline methods.}
    \label{fig:reg-plots-toussaint}
    \vspace{-0.5em}
\end{figure*}
\begin{figure}
\centering
        \includegraphics[width=4.8cm,alt={A scatter plot showing a cloud of points, where the x-axis represents the desired Rhythm Complexity attributes and the y-axis represents the Rhythm Complexity attributes of the corresponding sequences decoded by the LC-Diff model. A best-fit line illustrates the linear regression between them.}]{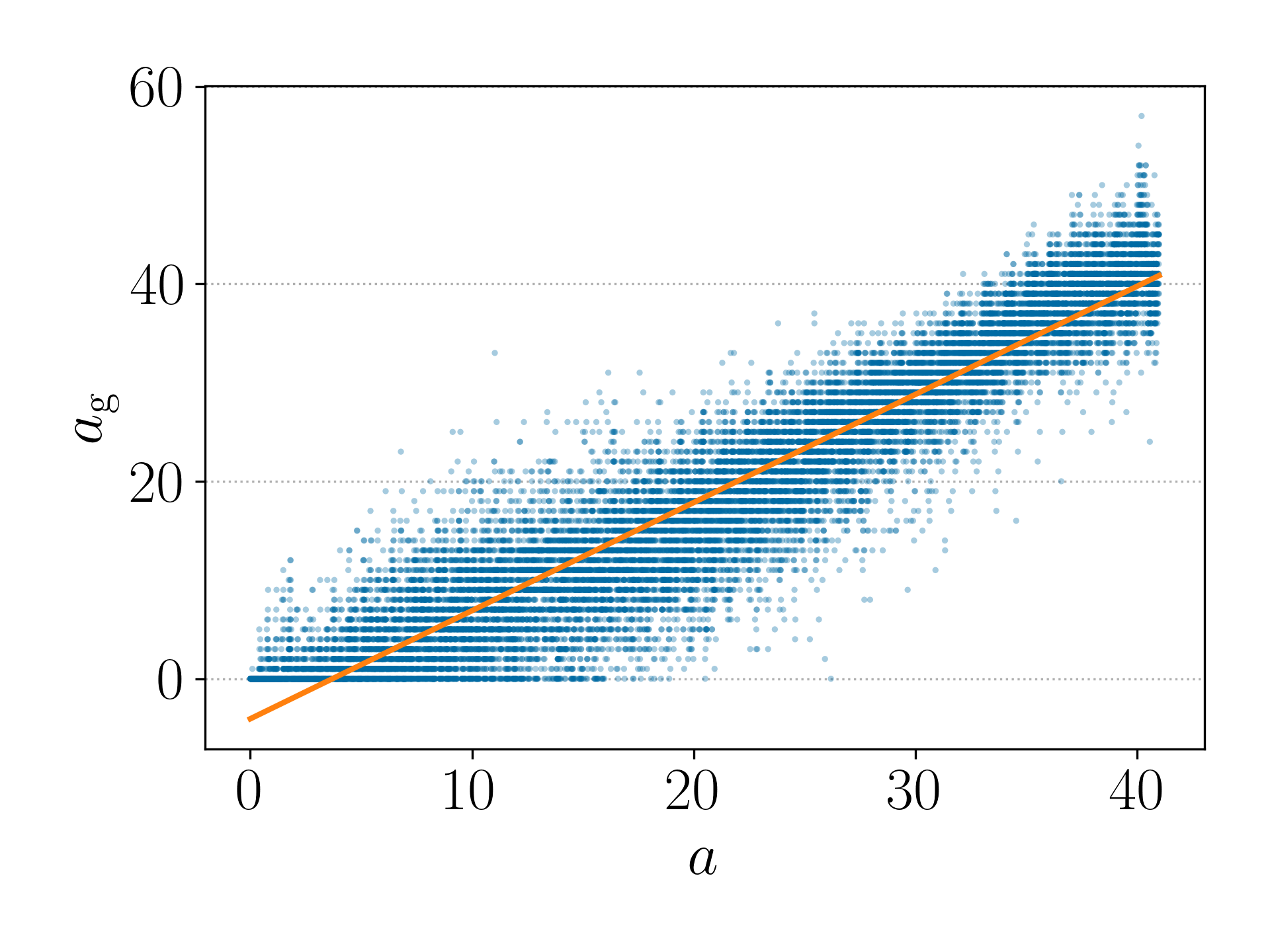}
    \vspace{-1em}
    \caption{Regression plot comparing target and decoded Rhythm Complexity attributes using LC-Diff.}
    \label{fig:reg-plots-toussaint-lc-diff}
    \vspace{-0.5em}
\end{figure}

\subsection{AR-VAE Baseline Methods}
\label{ssec:baselines}
For comparison, we consider AR-VAEs~\cite{mezzaLatentRhythmComplexity2023, patiAttributebasedRegularizationLatent2021} with the same architecture as the unconditional model described in Section~\ref{ssec:unconditional_model}. AR-VAEs incorporate regularization during training by means of a supervised multi-task learning approach, with the goal of encoding the attribute $a$ in the $i$-th dimension $z_i$ of their latent spaces. This is achieved by including an AR loss term in \eqref{eq:vae_objective}
\begin{equation}
\label{eq:ar_vae_objective} 
    \mathcal{L}_\text{AR-VAE} =  \mathcal{L}_\text{VAE} + \gamma \mathcal{L}_\text{AR},
\end{equation}
where $\gamma \ge 0$ is a tunable hyperparameter controlling the strength of the regularization.

Mezza et al.~\cite{mezzaLatentRhythmComplexity2023} propose the use of
\begin{align}
\label{eq:beta_vae_default_attr_reg_term}
    \mathcal{L}_{\text{AR}}^{\text{NM}} =  \operatorname{MAE}(z_i,\, \tilde{a}),
\end{align}
where $\operatorname{MAE}(\cdot,\cdot)$ denotes the mean absolute error, and $\tilde{a}$ is the z-score of $a$.

Pati and Lerch~\cite{patiAttributebasedRegularizationLatent2021} introduced a regularization term that enforces a monotonic relationship between $a$ and $z_i$, i.e.,
\begin{align}
\label{eq:beta_vae_sign_attr_reg_term}
    \mathcal{L}_{\text{AR}}^{\text{P\&L}} =  \operatorname{MAE} \left( \tanh(\delta\, \mathbf{D}_\mathrm{z}),\,  \operatorname{sign}(\mathbf{D}_\mathrm{a}) \right),
\end{align}
where $\mathbf{D}_\mathrm{z}$ and $\mathbf{D}_\mathrm{a}$ are pairwise distance matrices between $z_i$ and $a$ of all samples in a batch, respectively, and $\delta>0$ is a tunable hyperparameter. 
As in~\cite{patiAttributebasedRegularizationLatent2021}, we set $\gamma=1$ and $\delta=10$. The remaining training details are the same as in Section~\ref{ssec:unconditional_model}. For brevity, we will later refer to the former AR method as ``NM'' and to the latter as ``P\&L.''

\subsection{LC-VAE Baseline Methods}

Similarly to Tian and Engel~\cite{tian2019latent}, we implement LC through a conditional VAE (cVAE) trained on the representations of the base unconditional model (Section~\ref{ssec:unconditional_model}). 
The cVAE encoder consists of four linear layers with ReLU activations, followed by two Gating Mixing Layers (GML) that parameterize the innermost latent distribution. The decoder mirrors the encoder with four linear layers with ReLU activations, followed by an output GML. Except for using $2048$ units in the fully-connected layers and $M'=128$ latent variables, the cVAE architecture is the same as in~\cite{tian2019latent}.

Let $\bm{z}\in\mathbb{R}^M$ be the latent representations of the unconditional model, $\bm{z}_\mathrm{c} \in \mathbb{R}^{M'}$ be the latent representations of the cVAE, and $a\in\mathbb{R}$ the sequence attribute. The authors of~\cite{tian2019latent} considered binary labels and one-hot vectors were thus concatenated with $\mathbf{z}$ and $\mathbf{z}_\mathrm{c}$. Instead, we deal with continuous attributes.
We implement two cVAE variants that differ in how $a$ is fed into the networks. 
In the first variant, later referred to as LC-VAE-A,
we feed $\tilde{\bm{z}}=[\bm{z}^T, a]^T$ to the encoder, and $\tilde{\bm{z}}_\mathrm{c}=[\bm{z}_\mathrm{c}^T, a]^T$ to the decoder. In the second variant, named LC-VAE-SE, we concatenate $\bm{z}$ and $\bm{z}_\mathrm{c}$, respectively, with the attribute SE, i.e., $\tilde{\bm{z}}=[\bm{z}^T, \Gamma(a)]^T$ and $\tilde{\bm{z}}_\mathrm{c}=[\bm{z}_\mathrm{c}^T, \Gamma(a)]^T$.

\begin{figure*}
\centering
    \begin{subfigure}{0.32\linewidth}
        \centering
        \includegraphics[width=\linewidth,alt={Second of the three piano roll visualization of a melody generated by the LC-Diff model by specifying a conditioning value of 6 for the Contour attribute. The piano roll covers 4 bars, with each time unit representing 2 quarter notes, and spans 5 octaves from C1 to C6. This melody, generated with a Contour value of 5.85, shows a characteristic up-and-down melodic movement that reflects the specified Contour parameter.}]{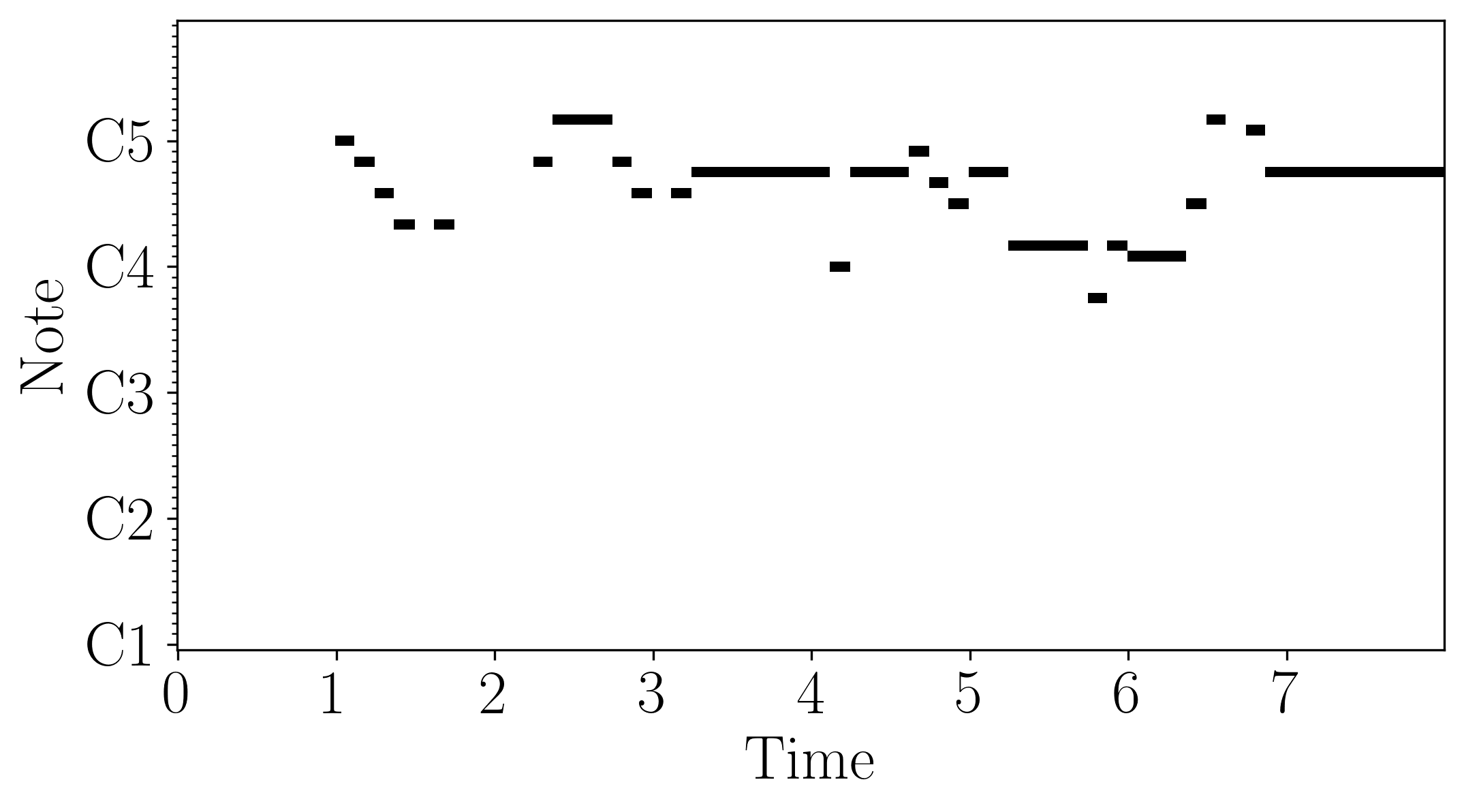}
        \caption{$a=1.0 \rightarrow a_\mathrm{g}=1.03$}
    \end{subfigure}\hfill
    \begin{subfigure}{0.32\linewidth}
        \centering
        \includegraphics[width=\linewidth,alt={Third of the three piano roll visualization of a melody generated by the LC-Diff model by specifying a conditioning value of 6 for the Contour attribute. The piano roll covers 4 bars, with each time unit representing 2 quarter notes, and spans 5 octaves from C1 to C6. This melody, generated with a Contour value of 5.49, shows a characteristic up-and-down melodic movement that reflects the specified Contour parameter.}]{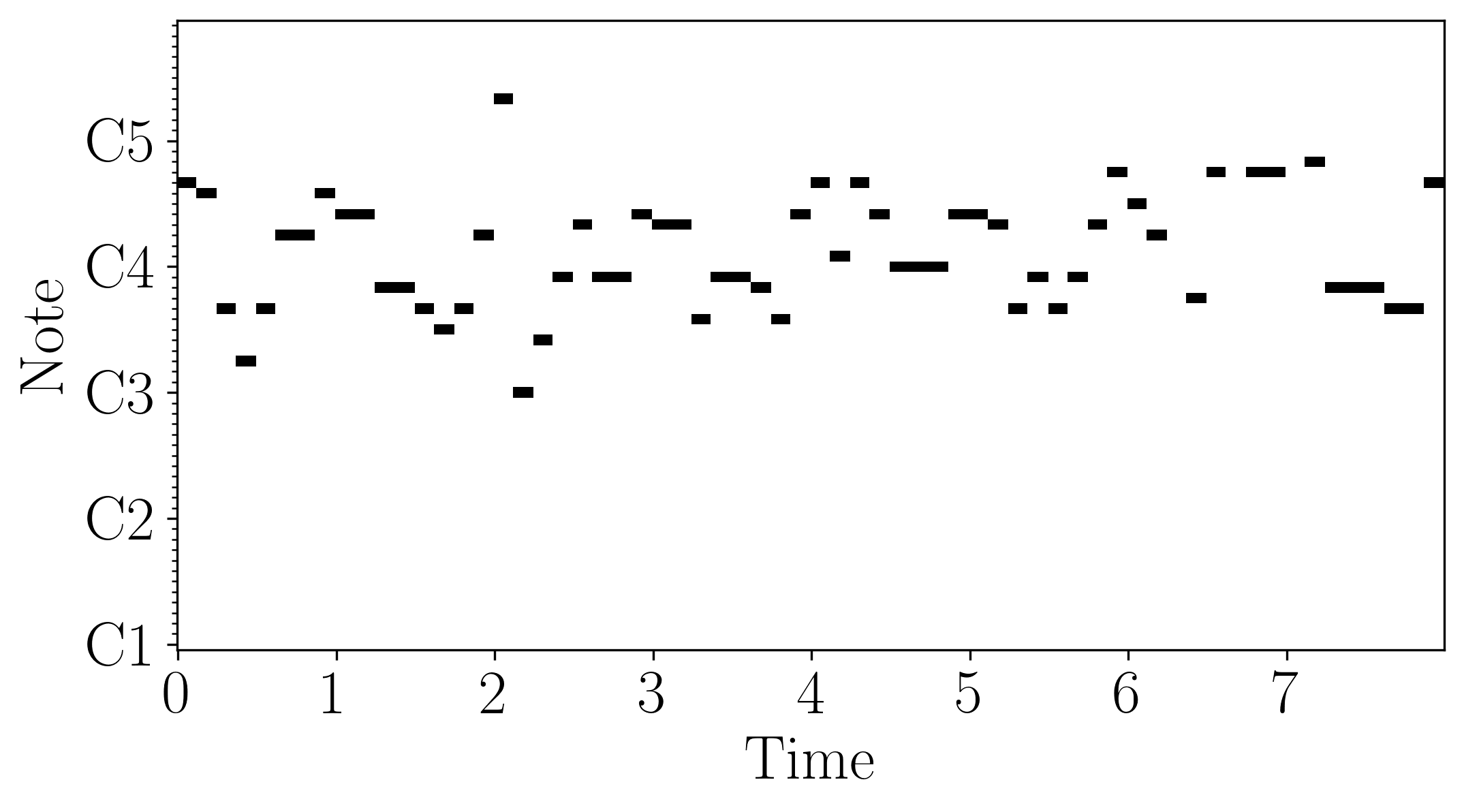}
        \caption{$a=3.0 \rightarrow a_\mathrm{g}=2.95$}
    \end{subfigure}\hfill
    \begin{subfigure}{0.32\linewidth}
        \centering
        \includegraphics[width=\linewidth,alt={First of the three piano roll visualization of a melody generated by the LC-Diff model by specifying a conditioning value of 6 for the Contour attribute. The piano roll covers 4 bars, with each time unit representing 2 quarter notes, and spans 5 octaves from C1 to C6. This melody, generated with a Contour value of 6.19, shows a characteristic up-and-down melodic movement that reflects the specified Contour parameter.}]{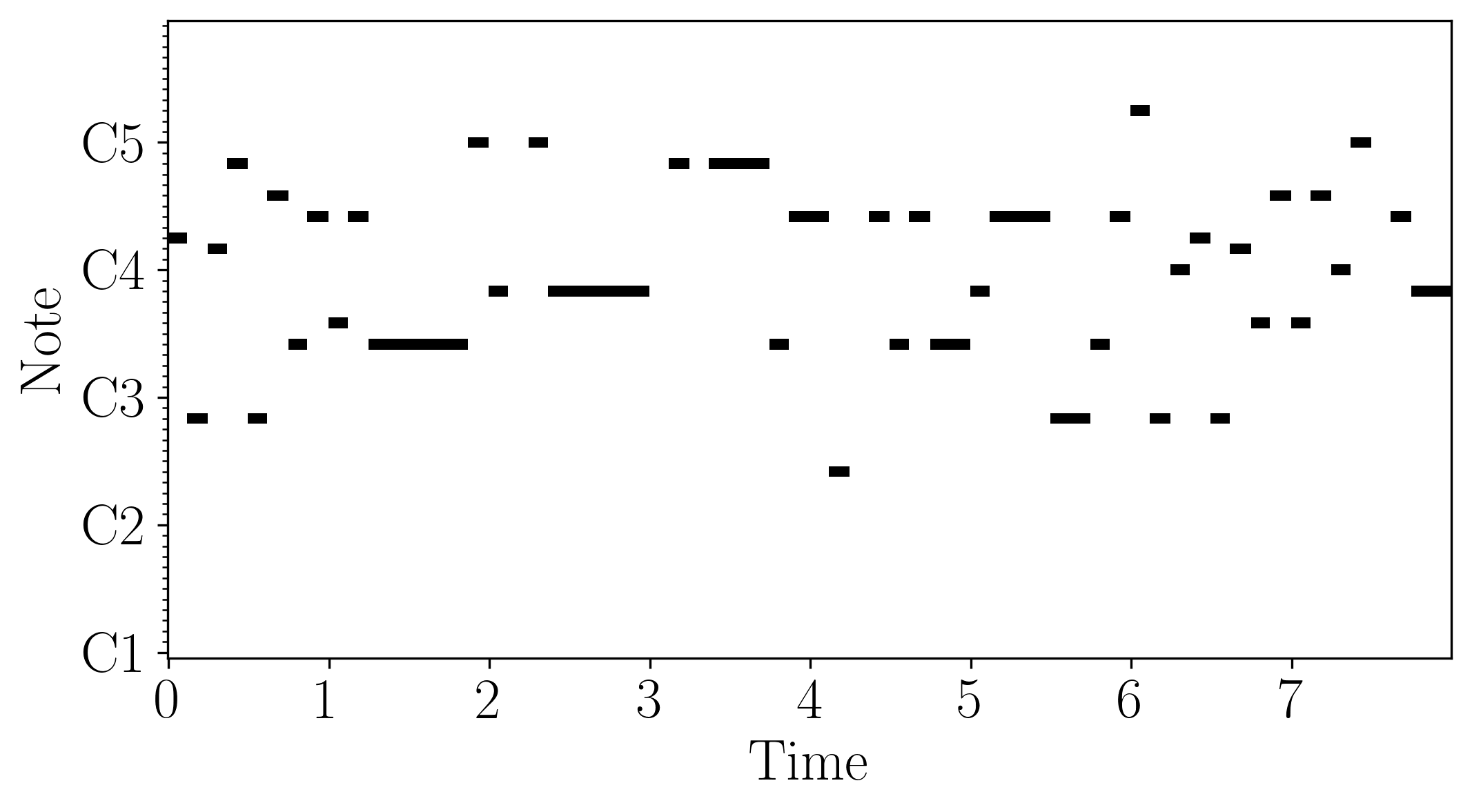}
        \caption{$a=6.0 \rightarrow a_\mathrm{g}=6.19$}
    \end{subfigure}\\
    \caption{Examples of MIDI files generated by controlling the Contour attribute with LC-Diff.}
    \label{fig:contour_examples}
\end{figure*}

\begin{figure*}
\centering
    \begin{subfigure}{0.32\linewidth}
        \centering
        \includegraphics[width=\linewidth,alt={First of the three piano roll visualization of a melody generated by the LC-Diff model by specifying a conditioning value of 6 for the Contour attribute. The piano roll covers 4 bars, with each time unit representing 2 quarter notes, and spans 5 octaves from C1 to C6. This melody, generated with a Contour value of 6.19, shows a characteristic up-and-down melodic movement that reflects the specified Contour parameter.}]{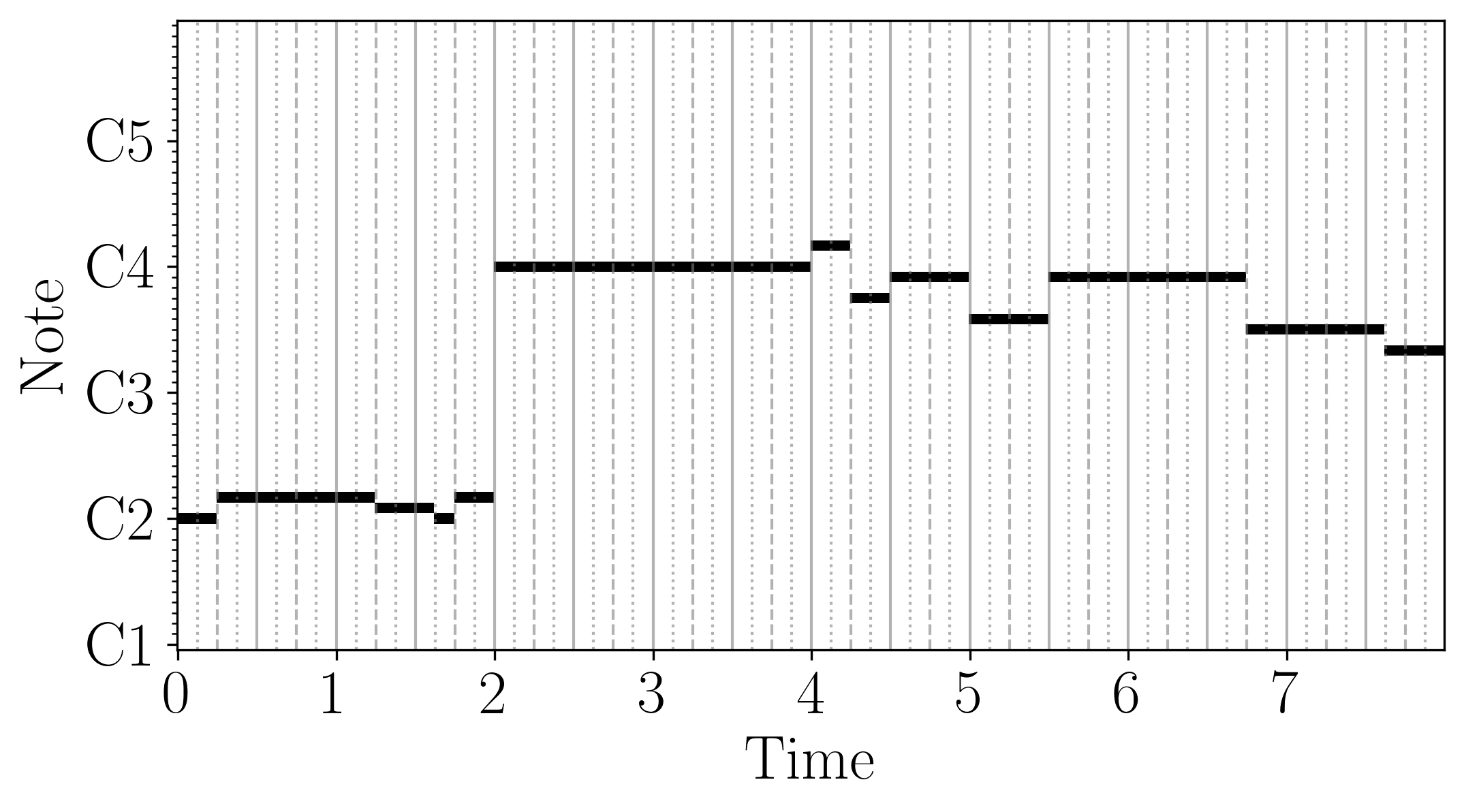}
        \caption{$a=0 \rightarrow a_\mathrm{g}=0$}
    \end{subfigure}\hfill
    \begin{subfigure}{0.32\linewidth}
        \centering
        \includegraphics[width=\linewidth,alt={Second of the three piano roll visualization of a melody generated by the LC-Diff model by specifying a conditioning value of 6 for the Contour attribute. The piano roll covers 4 bars, with each time unit representing 2 quarter notes, and spans 5 octaves from C1 to C6. This melody, generated with a Contour value of 5.85, shows a characteristic up-and-down melodic movement that reflects the specified Contour parameter.}]{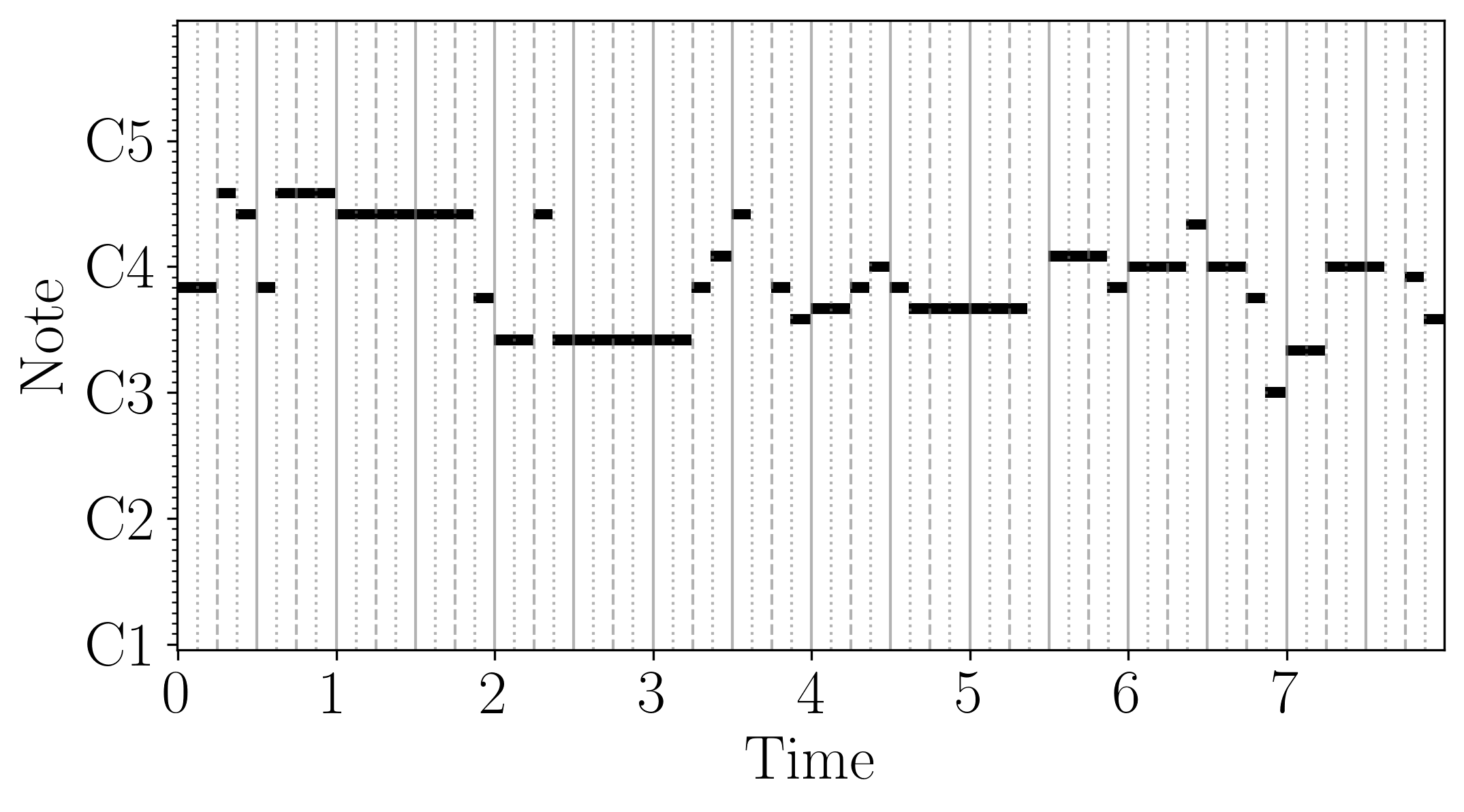}
        \caption{$a=15 \rightarrow a_\mathrm{g}=15$}
    \end{subfigure}\hfill
    \begin{subfigure}{0.32\linewidth}
        \centering
        \includegraphics[width=\linewidth,alt={Third of the three piano roll visualization of a melody generated by the LC-Diff model by specifying a conditioning value of 6 for the Contour attribute. The piano roll covers 4 bars, with each time unit representing 2 quarter notes, and spans 5 octaves from C1 to C6. This melody, generated with a Contour value of 5.49, shows a characteristic up-and-down melodic movement that reflects the specified Contour parameter.}]{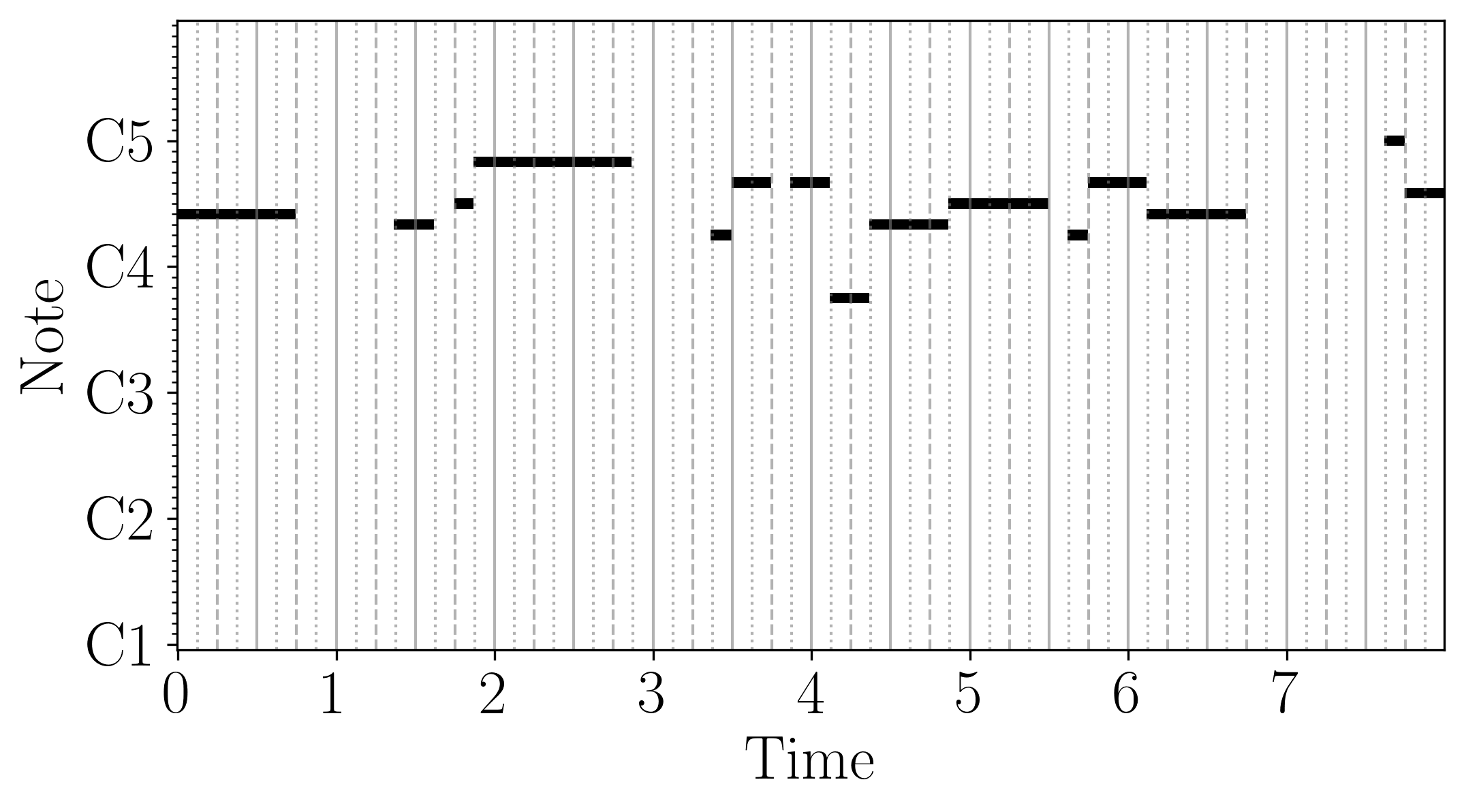}
        \caption{$a=33 \rightarrow a_\mathrm{g}=33$}
    \end{subfigure}\\
    \caption{Examples of MIDI files generated by controlling Rhythm Complexity with LC-Diff. Quarter notes are indicated by solid vertical lines; odd pulses (strong) are indicated by dashed lines; even pulses (weak) are indicated by dotted lines. }
    \label{fig:toussaint_examples}
\end{figure*}

\section{Results}\label{sec:results}

 
\subsection{Attribute-Controlled Generation}
\label{ssec:melody_generation}

To evaluate the controllability of the generative models under scrutiny, we sample the target attributes uniformly in the range of zero to the $99$th percentile of the attribute distribution of the sequences in the test set.\footnote{Limiting the range to the $99$th percentile is meant to exclude those sequences with abnormally high attribute values. We argue that these sequences are spurious, and we attribute their existence to the choice, borrowed from \cite{robertsHierarchicalLatentVector2018},
of extracting melodies by na{\"i}vely picking the highest note at any given time.}
These equally-spaced values, which we refer to as \textit{target} attributes, are fed to the respective conditioning network of LC-Diff, suitably transformed and plugged into the regularized dimension $z_i$ of the AR-VAEs, and concatenated to the input vector of the LC-VAE decoder networks.

Table \ref{tab:pearson_correlation} lists the Pearson Correlation Coefficients (PCC) between the target attributes and those computed from the generated sequences (the higher, the better).
LC-Diff consistently outperforms the two AR-VAEs (NM and P\&L) and LC-VAEs (both with and without SE) for all attributes considered. Notably, LC-Diff is the only method among those considered in the present study to yield correlation scores higher than $80\%$ across the board. 


As for Contour, LC-Diff achieves a PCC of $85.60\%$, outperforming the next-best model, NM, by over $22\%$. The difference is less pronounced for Note Density, where LC-Diff ($98.56\%$) improves upon the second-best model by just $1\%$. Nonetheless, LC-VAE-A and LC-VAE-SE already achieve $97.56\%$ and $97.33\%$, respectively, suggesting that constraining the generative model is very effective compared to AR methods when it comes to rendering the desired number of notes.
LC-Diff also demonstrates significant improvements in Pitch Range and Rhythm Complexity. For Pitch Range, it achieves a PCC of $80.97\%$, exceeding NM ($46.70\%$) by $34.27\%$, while NM itself outperforms LC-VAEs by approximately $10\%$. For Rhythm Complexity, LC-Diff achieves a remarkable $94.93\%$, surpassing LC-VAE-SE ($52\%$) by $42.93\%$.

Concerning AR models, while NM directly encodes the (standardized) distribution onto the $i$th dimension of the latent space, there is no a priori way to know the monotonic relationship learned using the P\&L regularization in \eqref{eq:beta_vae_sign_attr_reg_term}. This explains the near-zero correlation observed for Pitch Range, and, in general, the overall lower PCC.

Figures~\ref{fig:reg-plots-contour} through~\ref{fig:reg-plots-toussaint-lc-diff} show the regression plots of Contour and Rhythm Complexity. Figure~\ref{fig:reg-plots-contour} and Figure~\ref{fig:reg-plots-contour-lc-diff} illustrate the case of a continuous distribution, while Figure~\ref{fig:reg-plots-toussaint} and Figure~\ref{fig:reg-plots-toussaint-lc-diff} exemplify a case where the attribute takes on integer values.
Across both attributes, LC-Diff is characterized by a 
lower spread and a clear linear trend.
In Figure~\ref{fig:reg-plots-contour}, all baseline models show a tendency to produce excessively high contour values, whereas LC-Diff (Figure~\ref{fig:reg-plots-contour-lc-diff}) appears to mitigate the issue.
Likewise, Figure~\ref{fig:reg-plots-toussaint} reveals that all models but LC-Diff (Figure~\ref{fig:reg-plots-toussaint-lc-diff}) tend to fail when the target Complexity values are low.

\subsection{Data Fidelity}
\label{ssec:fmd}
To evaluate the quality of the generated sequences, we use the Fr{\'e}chet Music Distance (FMD)~\cite{retkowski2024frechet}, a metric that
extends the family of Fr{\'e}chet Inception Distance~\cite{heusel2017FID} and Fr{\'e}chet Audio Distance~\cite{Kilgour2019FAD} to the symbolic music domain. 
FMD was computed between $22,\!016$ melodies from the held-out test set and an equal number of generated sequences.
To prevent the FMD from measuring a spurious divergence from the real attribute distribution, we condition the generation on the attributes of the reference sequences, rather than using evenly-spaced control values as in Section~\ref{ssec:melody_generation}.
By conditioning with attributes measured from the test set, indeed, we aim to simultaneously compare the fidelity of generated sequences and how well they conform to the desired attribute distribution.

Table~\ref{tab:frechet_music_distance} reports the results obtained using CLaMP~2 MIDI embeddings \cite{wu2025clamp2multimodalmusic} (the lower, the better). 
For comparison, we report the FMD between the reference test set and the output of the unconditional VAE (see Section~\ref{ssec:unconditional_model}) obtained by decoding $22,\!016$ samples from $\mathcal{N}(\mathbf{0}, \mathbf{I})$.

The results presented in Table \ref{tab:frechet_music_distance} demonstrate that the proposed LC-Diff model consistently achieves the lowest FMD values across most attributes, indicating superior performance in generating samples that aligns more closely with the statistical properties of real sequences. Notably, LC-Diff outperforms all baselines in Contour ($19.299$), Note Density ($20.559$), and Rhythm Complexity ($17.51$), significantly improving over both AR-VAEs and LC-VAEs. While LC-VAE-A achieves the best Pitch Range score ($30.257$), LC-Diff remains competitive ($31.695$).

Overall, all LC methods outperform the unconditional base model ($41.44$), showing that introducing post-hoc control leads to more consistent and structured music generation, with better alignment to the desired attributes.

Finally, Figure~\ref{fig:contour_examples} and Figure~\ref{fig:toussaint_examples} illustrate the potential diversity in the generated samples produced by LC-Diff when conditioned on low, medium, and high values of Contour and Rhythm Complexity, respectively.



\section{Conclusions}\label{sec:conclusions}
In this paper, we have explored latent diffusion through the lens of Latent Constraints (LC), demonstrating the efficacy of DDIMs as plug-and-play conditioning modules for symbolic music generation. By keeping the base generative model fixed, we trained diffusion-based LC models (LC-Diff) capable of controlling a range of non-differentiable and continuous musical attributes, including contour, note density, pitch range, and rhythm complexity.
Our empirical evaluations reveal that LC-Diff significantly outperforms attribute-regularized VAEs and cVAE-based LC methods in terms of both fidelity and controllability, with absolute improvements of up to $12.65$ in Fréchet Music Distance and $43\%$ in correlation between desired and generated attributes.
These results highlight the potential of denoising as a powerful tool for ad hoc fader-like control over multiple musical attributes along continuous axes, effectively transforming a pre-trained unconditional model into a controllable music generation system depending on the user's needs.
Future work will focus on expanding the library of LC-Diff models to include a wider range of musical attributes and exploring the integration of user interfaces for real-time control.
Future experiments could also explore attribute-controlled input transformations by applying forward diffusion to encoded representations, rather than drawing noise samples from the standard normal prior.
Furthermore, we aim to investigate the potential for LC of other generative techniques, such as flow matching and consistency models.

\bibliography{ISMIRtemplate}

\end{document}